\documentclass[12pt,journal,draftcls,letterpaper,onecolumn]{IEEEtran}
\usepackage{epsfig}
\usepackage{graphicx}
\usepackage{amsmath}
\usepackage{amssymb}
\usepackage{amsfonts}
\usepackage[ruled,vlined]{algorithm2e}
\usepackage{algorithmic}
\usepackage{rotating}
\usepackage{multirow}
\usepackage{multicol}
\usepackage{blindtext}
\usepackage{subfigure}
\usepackage{float}
\usepackage{multicol}
\usepackage{lipsum}
\newtheorem{lemma}{Lemma}
\newtheorem{theorem}{Theorem}

\newtheorem{proposition}{Proposition}
\begin{document}

\title{Unmixing Incoherent Structures of Big Data by Randomized or Greedy Decomposition}

\author{Tianyi~Zhou,~Dacheng~Tao}


\maketitle

\begin{abstract}
Learning big data by matrix decomposition always suffers from expensive computation, mixing of complicated structures and noise. In this paper, we study more adaptive models and efficient algorithms that decompose a data matrix as the sum of semantic components with incoherent structures. We firstly introduce ``GO decomposition (GoDec)'', an alternating projection method estimating the low-rank part $L$ and the sparse part $S$ from data matrix $X=L+S+G$ corrupted by noise $G$. Two acceleration strategies are proposed to obtain scalable unmixing algorithm on big data: 1) Bilateral random projection (BRP) is developed to speed up the update of $L$ in GoDec by a closed-form built from left and right random projections of $X-S$ in lower dimensions; 2) Greedy bilateral (GreB) paradigm updates the left and right factors of $L$ in a mutually adaptive and greedy incremental manner, and achieve significant improvement in both time and sample complexities. Then we proposes three nontrivial variants of GoDec that generalizes GoDec to more general data type and whose fast algorithms can be derived from the two strategies: 1) for motion segmentation, we further decompose the sparse $S$ (moving objects) as the sum of multiple row-sparse matrices, each of which is a low-rank matrix after specific geometric transformation sequence and defines a motion shared by multiple objects; 2) for multi-label learning, we further decompose the low-rank $L$ into subcomponents with separable subspaces, each corresponds to the mapping a single label in feature space. Then the prediction can be effectively conducted by group lasso on the subspace ensemble; 3) for estimating scoring functions of each user in recommendation system, we further decompose the low-rank $L$ as $WZ^T$, where the rows of $W$ is the linear scoring functions and the rows of $Z$ are the items represented by available features. Empirical studies show the efficiency, robustness and effectiveness of the proposed methods in real applications.
\end{abstract}

\IEEEpeerreviewmaketitle

\section{Introduction}

\IEEEPARstart{C}{omplex} data is usually generated by mixing several components of different structures. These structures are often compressible, and are able to provide semantic interpretations of the data content. In addition, they can reveal the difference and similarity among data samples, and thus produce robust features playing vital roles in supervised or unsupervised learning tasks. Two types of structures have drawn lots of research attentions in recent years: 1) in compressed sensing \cite{DonohoCS, CandesT06}, a sparse signal can be exactly recovered from its linear measurements at a rate significant below the Nyquist rate, in sparse coding \cite{KSVD, MOD, EfficientSparseCoding}, an over-complete dictionary leads to sparse representations for dense signals of the same type; 2) in matrix completion \cite{MatrixCompletion, MatrixCompletionEC2, OptSpace, YeAccelerateNuclearNorm, SVP}, a low-rank matrix can be precisely rebuilt from a small portion of its entries by restricting the rows (samples) to lie in a subspace. In dimension reduction \cite{PCA, FisherLDA, LLE, ISOMAP, DLAtkde, MEN}, low-rank structure \cite{YeLowrank} has been broadly leveraged for exploring the geometry of point cloud. Although sparse and low-rank structures have been studied separately by a great number of researchers for years, the linear combination of them or their extensions is rarely explored until recently \cite{RankInco, RobustPCA, RobustMD, YeLowrankSparse}. Intuitively, fitting data with either sparse or low-rank structure is mature technique but is inevitably restricted by the limited data types they can model, while recent study shows that the linear mixture of them is more expressive in modeling complex data from different applications.

A motivating example is robust PCA \cite{RobustPCA} (RPCA), which decomposes the data matrix $X$ as $L+S$. The low-rank part $L$ summarizes a subspace that is shared by all the samples and thus reveals the global smoothness, while the sparse part $S$ captures the individual differences or abrupt changes among samples. A direct application of robust PCA is separating the sparse moving objects from the low-rank background in video sequence. Another interesting example is morphological component analysis (MCA) \cite{MCA}, which decompose the data into two parts that have sparse representations on two incoherent over-complete dictionaries, i.e., the first part has a very non-sparse representation on the dictionary of the second part, and vise versa. This requirement suggests that the two parts are separable on their sparse representations. Note that both RPCA and MCA can only work on data whose two building parts are incoherent, i.e., the content of one part cannot be moved to the other part without changing either of their structures (low-rank, sparse, dictionary, etc.). This incoherence condition could be viewed as a general extension of the statistical independence supporting independent component analysis (ICA) \cite{ICA, ICA1} blindly separating non-Gaussian source signals. It leads to the identifiability of the structures in theory, and is demonstrated to be fulfilled on a wide class of real data.

However, new challenges arises when many recent studies tend to focus on big data with complex structures. Firstly, existing algorithms are computationally prohibitive to processing these data. For instance, the update of low-rank part in RPCA and in its extensions invoke a full singular value decomposition (SVD) per iterate, while MCA requires challenging $\ell_0$ or $\ell_1$ minimization per sample/feature and previously achieved incoherent dictionaries/transform operators encouraging sparse representations. Thus they suffer from a dramatic growth in time complexity when either feature dimensions or data samples increase. In previous methods, the structured information such as low-rank and sparse properties are always achieved at the price of time-consuming optimization, but are rarely leveraged for the purpose of improving the scalability. Recent progresses in randomized approximation and rank-revealing algorithms shed some light on the speedup of the robust PCA typed algorithms: the subspace of the low-rank part can be estimated from random sampling of its columns/rows or projections of its columns/rows on a random ensemble with bounded precision \cite{RandomSVD, Stream, FJLT}. However, straightforward invoking this technique in RPCA problem needs to apply it to the updated residual matrix per iterate and thus may lead to costly computation. Besides, determining the rank of the low-rank part is not a trivial problem in practice.

Secondly, the simple low-rank, sparse and sparse representation assumptions cannot fully capture the sophisticated relation, individuality and sparsity of data samples with complex structures. While low-rank structure summarizes a global linear relationship between data points, the nonlinear relationship, local geometry and correlated functions are more common in big data and more expressive for a much wider class of structures. Moreover, the sparse matrix is simply explained by random noises on random positions in the past, but current studies reveal that it may have rich structured information that could be the central interests of various applications. For instance, the sparse motions captured by RPCA on video sequence data includes immense unexplored information favored by object tracking and behavior analysis. Furthermore, although the sparse representation is more general than sparse features, its quality largely relies on whether the given dictionary or transform operator fits the nature of data well. But this is difficult to evaluate when the data is of large volume and in general type.


Thirdly, two building parts are not sufficient to cover all the mixtures of incoherent structures in big data. One the one hand, dense noise is an extra component that has to be separated from the low-rank and sparse parts in many cases where the exact decomposition $X=L+S$ does not hold. This noisy assumption has been considered in stable PCP \cite{SPCP}, DRMF \cite{DRMF} and other theoretical studies \cite{RobustMD}, and its robustness and adaptiveness to a broad class of data has also been verified. But efficient algorithm for the noisy model lacks. On the other hand, further decomposing the low-rank or sparse part to multiple distinguishable sub-components is potential to tell locally spatial or temporal relations within each identifiable structure and differences between them, which usually play pivot roles in supervised and unsupervised learning tasks. Although it appeals to be a natural extension to the two-part model in RPCA, how to formulate a proper decomposition model for learning problems and develop a practical algorithm are challenging.


\subsection{Main Contributions}

We start this paper by studying a novel low-rank and sparse matrix decomposition model ``GO decomposition (GoDec)'' \cite{GoDec} $X=L+S+G$, which takes an extra dense noisy part $G$ into account and casts the decomposition into alternating optimization of low-rank $L$ and sparse $S$. In order to overcome the computational burden caused by the large volume of data, we propose two acceleration strategies in designing the decomposition algorithms: the first is ``bilateral random projection (BRP)'' \cite{BRP} based fast low-rank approximation that results in a randomized update of the low-rank part or its nonlinear variant, this technique is based on recently developed random matrix theories that show a few random projections of a matrix is able to reveal its associated principle subspace \cite{RandomProjection, RandomSVD, Stream, FJLT}; the other is a Frank-Wolfe typed optimization scheme called ``greedy bilateral (GreB)'' paradigm \cite{GreB} that updates the left and right factors of the low-rank matrix variable in a mutually adaptive and greedy incremental manner. We show the two strategies generates considerably scalable algorithms for low-rank and sparse matrix decomposition. Moreover, both strategies have provable performance guarantee given by rigorous theoretical analysis (Appendix I and II).

In order to deal with the complicated structures that cannot be captured by the sum mixture of low-rank and sparse matrices, we proposes three variants of GoDec more expressive and general for learning from big data.

The first variant ``shifted subspace tracking (SST)'' \cite{SST} is developed for motion segmentation \cite{motionICCV, crowdCVPR, DetectFree, TraPF, HiePar} from raw pixels of video sequence. SST further analyzes the unexplored rich structure of the sparse part $S$ of GoDec, which could be seem as a sum mixture of several motions with distinct appearance and trajectories. SST unifies detection, tracking and segmenting multiple motions from complex scenes in a simple matrix factorization model.

The second variant ``multi-label subspace ensemble (MSE)'' \cite{MSE} extends the low-rank part $L$ of GoDec to the sum of multiple low-rank matrices defined by distinguishable but correlated subspaces. MSE provides a novel insight into the multi-label learning (ML) problem \cite{LearningFML, MLreview1, MLreview2, ReverseML, HOMER}. It addresses this problem by jointly learning inverse mappings that map each label to the feature space as a subspace, and formulating the prediction as finding the group sparse representation \cite{GroupLasso} of a given sample on the ensemble of subspaces. There are only $k$ subspaces needed to be learned, and the label correlations are fully used via considering correlation among subspaces.

The third variant ``linear functional GoDec (LinGoDec)'' learns scoring functions of users from their ratings matrix $X$ and features of scored items $Z$. It extends the low-rank part $L$ of GoDec to $WZ^T$, where $W$ represents the linear functions and is constrained to be low-rank, while the rows of $Z$ contain the features of items in the training set. In addition, the sparse part $S$ is able to detect the advertising effects or anomaly of users' ratings on specific items. LinGoDec formulates the collaborative filtering problem as supervised learning, and thus avoids time-consuming completion of the whole matrix when only a new item's scores (a new row) are needed to be predicted.

The rest of this paper is organized as following: Section 2 introduces GoDec; Section 3 proposes the two acceleration strategies for processing large-scale data; Section 4 proposes the three variants of GoDec and their practical algorithms; Section 5 shows the experimental results of all the proposed algorithms on different application problems and justifies both the effectiveness and efficiency of them. The rows of all data matrices mentioned in this paper represents the samples and the columns denote the features.

\section{GO Decomposition: Unmixing Low-rank and Sparse Structures}


In RPCA \cite{RobustPCA}, PCP recovers $L$ and $S$ from $X$ by minimizing sum of the trace norm of $L$ and the $\ell_1$ norm of $S$. It can be proved that the solution to this convex relaxation is the exact recovery if $X=L+S$ indeed exists and $L$ and $S$ are sufficiently incoherent \cite{RankInco, RobustPCA}. That is, $L$ obeys the incoherence property and thus is not sparse, while $S$ has nonzero entries uniformly selected at random and thus is not low-rank. Popular optimization algorithms such as augmented Lagrangian multiplier, accelerated proximal gradient method and accelerated projected gradient method \cite{YeLowrankSparse} have been applied. But full SVD as a costly subroutine is required to be repeatedly invoked in any of them.

Despite the strong theoretical guarantee of robust PCA, the exact decomposition $X=L+S$ does not always hold for real data matrix $X$ due to extra noise and complicated structure of $S$ that does not following Bernoulli-Gaussian distribution. Thus a more adaptive model $X=L+S+G$ is preferred, where $L+S$ approximates $X$ and $G$ is the dense noise. We then study the approximated ``low-rank+sparse'' decomposition of a matrix $X$, i.e.,
\begin{equation}\label{E:GoDecMODEL}
X=L+S+G,{\rm rank}(L)\leq r,{\rm card}(S)\leq k,
\end{equation}
In this section, we develop ``Go Decomposition'' (GoDec) to estimate the low-rank part $L$ and the sparse part $S$ from $X$ by solving the following optimization problem,  which aims at minimizing the decomposition error:
\begin{equation}\label{E:ls_app}
\begin{array}{rl}
\min\limits_{L,S}&\|X-L-S\|_F^2\\
s.t.&{\rm rank}\left(L\right)\leq r,\\
&{\rm card}\left(S\right)\leq k.
\end{array}
\end{equation}

\subsection{Na\"{\i}ve GoDec}

We propose the na\"{\i}ve GoDec algorithm at first and will study how to achieve an highly accelerated version in the next section. The optimization problem of GoDec (\ref{E:ls_app}) can be solved by alternatively solving the following two subproblems until convergence:
\begin{equation}\label{E:raw_ls}
\left\{
  \begin{array}{ll}
    L_t=\arg\min\limits_{{\rm rank}\left(L\right)\leq r}\|X-L-S_{t-1}\|_F^2; \\
    S_t=\arg\min\limits_{{\rm card}\left(S\right)\leq k}\|X-L_t-S\|_F^2.
  \end{array}
\right.
\end{equation}
Although both subproblems (\ref{E:raw_ls}) have nonconvex constraints, their global solutions $L_t$ and $S_t$ exist. Let the SVD of a matrix $X$ be $U\Lambda V^T$ and $\lambda_i$ or $\lambda_i(X)$ stands for the $i^{th}$ largest singular value of $X$; $\mathcal P_\Omega\left(\cdot\right)$ is the projection of a matrix to an entry set $\Omega$.

In particular, the two subproblems in (\ref{E:raw_ls}) can be solved by updating $L_t$ via singular value hard thresholding of $X-S_{t-1}$ and updating $S_t$ via entry-wise hard thresholding of $X-L_t$, respectively, i.e.,
\begin{equation}\label{E:raw_lss}
\left\{
  \begin{array}{ll}
    L_t=\sum\limits_{i=1}^r\lambda_iU_iV_i^T, {\rm svd}\left(X-S_{t-1}\right)=U\Lambda V^T; \\
    S_t=\mathcal {P}_{\Omega}\left(X-L_t\right), \Omega:\left|\left(X-L_t\right)_{{i,j}\in{\Omega}}\right|\neq0 \\ ~~~~{\rm~and~} \geq \left|\left(X-L_t\right)_{{i,j}\in{\overline{\Omega}}}\right|, |\Omega|\leq k.
  \end{array}
\right.
\end{equation}
The main computation in the na\"{\i}ve GoDec algorithm (\ref{E:raw_lss}) is the SVD of $X-S_{t-1}$ in the updating $L_t$ sequence. SVD requires $\min\left(mn^2,m^2n\right)$ flops, so it is impractical when $X$ is of large size, and more efficient algorithm is needed to be developed later.

GoDec alternatively assigns the $r$-rank approximation of $X-S$ to $L$ and assigns the sparse approximation with cardinality $k$ of $X-L$ to $S$. The updating of $L$ is obtained via singular value hard thresholding of $X-S$, while the updating of $S$ is obtained via entry-wise hard thresholding \cite{IterativeHardThresh} of $X-L$. The term ``GO'' is owing to the similarities between $L$/$S$ in the GoDec iteration rounds and the two players in the game of go. 

Except the additional noisy part $G$ and faster speed, the direct constraints to the rank of $L$ and the cardinality $S$ also makes GoDec different from RPCA minimizing their convex polytopes. This makes the rank and cardinality controllable, which is preferred in practice. Because prior information of these two parameters can be applied and lots of computations might be saved. In addition, GoDec introduces an efficient matrix completion algorithm \cite{GoDec}, in which the cardinality constraint is replaced by a fixed support set. Convergence and robustness analysis of GoDec is given in Appendix I based on theory of alternating projection on two manifolds \cite{AlternatingP}. 


\section{Two Acceleration Strategies for Unmixing Incoherent Structures}

We firstly introduce the bilateral random projections (BRP) based low-rank approximation and its power scheme modification. BRP reduces the time consuming SVD in na\"{\i}ve GoDec to a closed-form approximation merely requiring small matrix multiplications. However, we need to invoke more expensive power scheme of BRP when the matrix spectrum does not have dramatic decreasing. Moreover, the rank needs to be estimated for saving unnecessary computations. Thus we propose greedy bilateral sketch (GreBske), which augments the matrix factors column/rows-wisely by selecting the best rank-one directions for approximation. It can adaptively determines the rank by stopping the augmenting when error is sufficiently small, and has accuracy closer to SVD.

\subsection{Bilateral Random Projection based Strategy}

\subsubsection{Low-rank approximation with closed form}

Given $r$ bilateral random projections (BRP) of an $m\times n$ dense matrix $X$ (w.l.o.g, $m\geq n$), i.e., $Y_1=XA_1$ and $Y_2=X^TA_2$, wherein $A_1\in\mathbb R^{n\times r}$ and $A_2\in\mathbb R^{m\times r}$ are random matrices,
\begin{equation}\label{E:lr_app}
L=Y_1\left(A_2^TY_1\right)^{-1}Y_2^T
\end{equation}
is a fast rank-$r$ approximation of $X$. The computation of $L$ includes an inverse of an $r\times r$ matrix and three matrix multiplications. Thus, for a dense $X$, $2mnr$ floating-point operations (flops) are required to obtain BRP, $r^2(2n+r)+mnr$ flops are required to obtain $L$. The computational cost is much less than SVD based approximation. 

In order to improve the approximation precision of $L$ in (\ref{E:lr_app}) when $A_1$ and $A_2$ are standard Gaussian matrices, we use the obtained right random projection $Y_1$ to build a better left projection matrix $A_2$, and use $Y_2$ to build a better $A_1$. In particular, after $Y_1=XA_1$, we update $A_2=Y_1$ and calculate the left random projection $Y_2=X^TA_2$, then we update $A_1=Y_2$ and calculate the right random projection $Y_1=XA_1$. A better low-rank approximation $L$ will be obtained if the new $Y_1$ and $Y_2$ are applied to (\ref{E:lr_app}). This improvement requires additional flops of $mnr$ in BRP calculation.

\subsubsection{Power scheme modification}

When singular values of $X$ decay slowly, (\ref{E:lr_app}) may perform poorly. We design a modification for this situation based on the power scheme \cite{PowerScheme}. In the power scheme modification, we instead calculate the BRP of a matrix $\tilde X=(XX^T)^qX$, whose singular values decay faster than $X$. In particular, $\lambda_i(\tilde X)={\lambda_i(\tilde X)}^{2q+1}$. Both $X$ and $\tilde X$ share the same singular vectors. The BRP of $\tilde X$ is:
\begin{equation}
Y_1=\tilde XA_1, Y_2=\tilde X^TA_2.
\end{equation}
According to (\ref{E:lr_app}), the BRP based $r$ rank approximation of $\tilde X$ is:
\begin{equation}
\tilde L=Y_1\left(A_2^TY_1\right)^{-1}Y_2^T.
\end{equation}
In order to obtain the approximation of $X$ with rank $r$, we calculate the QR decomposition of $Y_1$ and $Y_2$, i.e.,
\begin{equation}
Y_1=Q_1R_1, Y_2=Q_2R_2.
\end{equation}
The low-rank approximation of $X$ is then given by:
\begin{equation}\label{E:mlr_app}
L=\left(\tilde L\right)^{\frac{1}{2q+1}}=Q_1\left[R_1\left(A_2^TY_1\right)^{-1}R_2^T\right]^{\frac{1}{2q+1}}Q_2^T.
\end{equation}
The power scheme modification (\ref{E:mlr_app}) requires an inverse of an $r\times r$ matrix, an SVD of an $r\times r$ matrix and five matrix multiplications. Therefore, for dense $X$, $2(2q+1)mnr$ flops are required to obtain BRP, $r^2(m+n)$ flops are required to obtain the QR decompositions, $2r^2(n+2r)+mnr$ flops are required to obtain $L$. The power scheme modification reduces the error of (\ref{E:lr_app}) by increasing $q$. When the random matrices $A_1$ and $A_2$ are built from $Y_1$ and $Y_2$, $mnr$ additional flops are required in the BRP calculation. Thorough error bound analysis of BRP and its power scheme is given in Appendix II.

\subsubsection{Fast GoDec by Bilateral Random Projection}

Since BRP based low-rank approximation is near optimal and efficient, we replace SVD with BRP in na\"{\i}ve GoDec in order to significantly reduce the time cost.

We summarize GoDec using BRP based low-rank approximation (\ref{E:lr_app}) and power scheme modification (\ref{E:mlr_app}) in Algorithm 1. When $q=0$, For dense $X$, (\ref{E:lr_app}) is applied. Thus the QR decomposition of $Y_1$ and $Y_2$ in Algorithm 1 are not performed, and $L_t$ is updated as $L_t=Y_1\left(A_2^TY_1\right)^{-1}Y_2^T$. In this case, Algorithm \ref{alg:GoDec} requires $r^2\left(2n+r\right)+4mnr$ flops per iteration. When integer $q>0$, (\ref{E:mlr_app}) is applied and Algorithm 1 requires $r^2\left(m+3n+4r\right)+(4q+4)mnr$ flops per iteration.

\begin{algorithm}[htb]\label{alg:GoDec}\footnotesize
\SetAlgoLined
\KwIn{$X$, $r$, $k$, $\epsilon$, $q$}
\KwOut{$L$, $S$}
\textbf{Initialize} $L_0:=X$, $S_0:=\textbf{0}$, $t:=0$\;
\While{$\|X-L_t-S_t\|_F^2/\|X\|_F^2>\epsilon$}{
$t:=t+1$\;
$\tilde L=\left[\left(X-S_{t-1}\right)\left(X-S_{t-1}\right)^T\right]^q\left(X-S_{t-1}\right)$\;
$Y_1=\tilde LA_1$, $A_2=Y_1$\;
$Y_2=\tilde L^TY_1=Q_2R_2$, $Y_1=\tilde LY_2=Q_1R_1$\;
\textbf{If} ${\rm rank}\left(A_2^TY_1\right)<r$ \textbf{then} $r:={\rm rank}\left(A_2^TY_1\right)$, go to the first step; \textbf{end}\;
$L_t=Q_1\left[R_1\left(A_2^TY_1\right)^{-1}R_2^T\right]^{1/\left(2q+1\right)}Q_2^T$\;
$S_t=\mathcal {P}_{\Omega}\left(X-L_t\right)$, $\Omega$ is the nonzero subset of the first $k$ largest entries of $|X-L_t|$\;
}
\caption{GO Decomposition (GoDec) by BRP}
\end{algorithm}\normalsize

\subsection{Greedy Bilateral Factorization Strategy}

The major computation in na\"{\i}ve GoDec is the update of the low-rank part $L$, which requires at least a truncated SVD. Although the proposed randomized strategy provides a faster and SVD-free algorithm for GoDec, how to determine the rank of $L$ and the cardinality of $S$ is still an unsolved problem in real applications. In fact, these two parameters are not easy to determine and could lead to unstable solutions when estimated incorrectly. Noisy robust PCA methods such as stable PCP \cite{SPCP}, GoDec \cite{GoDec} and DRMF \cite{DRMF} usually suffer from this problem. Another shortcoming of the randomized strategy is that the time complexity is dominated by matrix multiplications, which could be computationally slow on high-dimensional data. 

In this part, we describe and analyze a general scheme called ``greedy bilateral (GreB)'' paradigm for solving optimizing low-rank matrix in mainstream problems. In GreB, the low-rank variable $L$ is modeled in a bilateral factorization form $UV$, where $U$ is a tall matrix and $V$ is a fat matrix. It starts from $U$ and $V$ respectively containing a very few (e.g., one) columns and rows, and optimizes them alternately. Their updates are based on observation that the object value is determined by the product $UV$ rather than individual $U$ or $V$. Thus we can choose a different pair $(U,V)$ producing the same $UV$ but computed faster than the one derived by alternating least squares like in IRLS-M \cite{IRLSM} and ALS \cite{ALS}. In GreB, the updates of $U$ and $V$ can be viewed as mutually adaptive update of the left and right sketches of the low-rank matrix. Such updates are repeated until the object convergence, then a few more columns (or rows) are concatenated to the obtained $U$ (or $V$), and the alternating updates are restarted on a higher rank. Here, the added columns (or rows) are selected in a greedy manner. Specifically, they are composed of the rank-$1$ column (or row) directions on which the object decreases fastest. GreB incrementally increases the rank until when $UV$ is adequately consistent with the observations.

\begin{algorithm}[htb]\label{alg:GreB}\footnotesize
\SetAlgoLined
\KwIn{Object function $f$; rank step size $\Delta r$; power $K$; tolerance $\tau$; observations of data matrix $X$}
\KwOut{low-rank matrix $UV$ and sparse $S$}
\textbf{Initialize} $V\in\mathbb R^{r_0\times n}$ (and $S$)\;
\While{residual error $\leq\tau$}{
\For{$k\leftarrow 1$ \KwTo $K$}{
Update $U$, $V$ and $S$ by alternating minimization rules, other faster $U$ and $V$ update rules can be applied if they produce equal $UV$\;
Greedy Bilateral Smoothing: sequentially compute (\ref{equ:GreBsmoA})\;
}
Calculate the top $\Delta r$ right singular vectors $v$ (or $\Delta r$-dimensional random projections) of $\partial f/\partial V$ (for GreBsmo compute (\ref{equ:GreBsmoD}));
Set $V:=[V;v]$\;
}
\caption{Greedy Bilateral (GreB) Paradigm}
\end{algorithm}\normalsize

GreB's greedy strategy avoids the failures brought by possible biased rank estimation. Moreover, greedy selecting optimization directions from $1$ to $r$ is faster than updating $r$ directions in all iterates like in LMaFit \cite{LMaFit} and \cite{GoDec}. In addition, the lower rank solution before each rank increment is invoked as the ``warm start'' of the next higher rank optimization and thus speed up convergence. Furthermore, its mutually adaptive updates of $U$ and $V$ yields a simple yet efficient SVD-free implementation. Under GreB paradigm, the overall time complexity of matrix completion is $\mathcal O(\max\{\|\Omega\|_0r^2,(m+n)r^3\})$ ($\Omega$-sampling set, $m\times n$-matrix size, $r$-rank), while the overall complexities of low-rank approximation and noisy robust PCA are $\mathcal O(mnr^2)$. An improvement on sample complexity can also be justified. An theoretical analysis of GreB solution convergence based on the result of GECO \cite{GECO} is given in Appendix III.

In the following, we present GreB by using it to derive a practical algorithm ``greedy bilateral smoothing (GreBsmo)'' for GoDec. It can also be directly applied to low-rank approximation and matrix completion []. We summarize general GreB paradigm in Algorithm \ref{alg:GreB}, and then present the detailed GreBsmo algorithm.

\subsubsection{Faster GoDec by Greedy Bilateral Smoothing}

In particular, we formulate GoDec by replacing $L$ with its bilateral factorization $L=UV$ and regularizing the $\ell_1$ norm of $S$'s entries:
\begin{equation}\label{equ:GreBsmo}
\begin{array}{ll}
&\min_{U,V,S}\|X-UV-S\|_F^2+\lambda\|{\rm vec}(S)\|_1\\
&{\rm s.t.}~~rank(U)=rank(V)\leq r.
\end{array}
\end{equation}
Note the $\ell_1$ regularization is a minor modification to the cardinality constraint in (\ref{E:ls_app}). It induces soft-thresholding in updating $S$, which is faster than sorting caused by cardinality constraint in GoDec and DRMF.

Alternately optimizing $U$, $V$ and $S$ in (\ref{equ:GreBsmo}) immediately yields the following updating rules:
\begin{equation}\label{equ:NGreBsmo}
\left\{
  \begin{array}{ll}
    U_k=\left(X-S_{k-1}\right)V_{k-1}^T\left(V_{k-1}V_{k-1}^T\right)^\dag, \\
    V_k=\left(U_k^TU_k\right)^\dag U_k^T\left(X-S_{k-1}\right), \\
    S_k=\mathcal S_\lambda\left(X-U_kV_k\right),
  \end{array}
\right.
\end{equation}
where $\mathcal S_\lambda$ is an element-wise soft thresholding operator with threshold $\lambda$ such that
\begin{equation}\label{equ:softt}
\mathcal S_\lambda X=\left\{{\rm sgn}\left(X_{ij}\right)\max\left(\left|X_{ij}\right|-\lambda,0\right):(i,j)\in[m]\times [n]\right\}.
\end{equation}
The same trick of replacing the $(U,V)$ pair with a faster computed one is applied and produce
\begin{equation}\label{equ:GreBsmoA}
\left\{
  \begin{array}{ll}
    U_k=Q, {\rm QR}\left(\left(X-S_{k-1}\right)V_{k-1}^T\right)=QR, \\
    V_k=Q^T\left(X-S_{k-1}\right), \\
    S_k=\mathcal S_\lambda\left(X-U_kV_k\right),
  \end{array}
\right.
\end{equation}
The above procedure can be performed in $3mnr_i+mr_i^2$ flops for $U\in\mathbb R^{m\times r_i}$ and $V\in\mathbb R^{r_i\times n}$.

In GreBsmo, (\ref{equ:GreBsmoA}) is iterated as a subroutine of GreB's greedy incremental paradigm. In particular, the updates in (\ref{equ:GreBsmoA}) are iterated for $K$ times or until the object converging, then $\Delta r$ rows are added into $V$ as the new directions for decreasing the object value. In order to achieve the fastest decreasing directions, we greedily select the added $\Delta r$ rows as the top $\Delta r$ right singular vectors of the partial derivative
\begin{equation}\label{equ:GreBsmoD}
\frac{\partial \|X-UV-S\|_F^2}{\partial V}=X-UV-S.
\end{equation}
We also allow to approximate row space of the singular vectors via random projections \cite{RandomSVD}. The selected $\Delta r$ rows maximize the magnitude of the above partial derivative and thus lead to the most rapid decreasing of the object value, a.k.a., the decomposition error. GreBsmo repeatedly increases the rank until a sufficiently small decomposition error is achieved. So the rank of the low-rank component is adaptively estimated in GreBsmo and does not relies on initial estimation.

\section{Three Variants of GoDec}

Although the two strategies successfully generate efficient low-rank and sparse decomposition capable to tackle large volume problem of big data, the complicated structures widely existing in big data cannot be always expressed by the sum of low-rank and sparse matrices and thus may still lead to the failure of RPCA typed models. Therefore, we address this problem by developing several GoDec's variants that unravel different combination of incoherent structures beyond low-rank and sparse matrices, where the two strategies can be still used to achieve scalable algorithms. 

\subsection{Shifted Subspace Tracking (SST) for Motion Segmentation}

SST decomposes $S$ of GoDec into the sum of several matrices, each of whose rows are generated by imposing a smooth geometric transformation sequence to the rows of a low-rank matrix. These rows store moving object in the same motion after aligning them across different frames, while the geometric transformation sequence defines the shared trajectories and deformations of those moving objects across frames. In the following, we develop an efficient randomized algorithm extracting the motions in sequel, where the low-rank matrix for each motion is updated by BRP, and the geometric transformation sequence is updated in a piece-wise linear approximation manner. 

We consider the problem of motion segmentation from the raw video data. Given a data matrix $X\in\mathbb R^{n\times p}$ that stores a video sequence of $n$ frames, each of which has $w\times h=p$ pixels and reshaped as a row vector in $X$, the goal of SST framework is to separate the motions of different object flows, recover both their low-rank patterns and geometric transformation sequences. This task is decomposed as two steps, background modeling that separates all the moving objects from the static background, and flow tracking that recovers the information of each motion. In this problem, $\cdot_i$ stands for the $i^{th}$ entry of a vector or the $i^{th}$ row of a matrix, while $\cdot_{i,j}$ signifies the entry at the $i^{th}$ row and the $j^{th}$ column of a matrix.

The first step can be accomplished by either GoDec or GreBsmo. After obtaining the sparse outliers $S$ storing multiple motions, SST treats the sparse matrix $S$ as the new data matrix $X$, and decomposes it as $X=\sum\nolimits_{i=1}^k\tilde L(i)+S+G$, wherein $\tilde L(i)$ denotes the $i^{th}$ motion, $S$ stands for the sparse outliers and $G$ stands for the Gaussian noise.

The motion segmentation in SST is based on an observation to the implicit structures of the sparse matrix $\tilde L(i)$. If the trajectory of the object flow $\tilde L(i)$ is known and each frame (row) in $\tilde L(i)$ is shifted to the position of a reference frame, due to the limited number of poses for the same object flow in different frames, it is reasonable to assume that the rows of the shifted $\tilde L(i)$ exist in a subspace. In other words, $\tilde L(i)$ after inverse geometric transformation is low-rank. Hence the sparse motion matrix $\tilde L(i)$ has the following structured representation
    \begin{equation}
    \tilde L(i)=
    \left[
      \begin{array}{c}
        L(i)_1\circ\tau(i)_1 \\
        \vdots \\
        L(i)_n\circ\tau(i)_n \\
      \end{array}
    \right]=L(i)\circ\tau(i).
    \end{equation}
The invertible transformation $\tau(i)_j: \mathbb R^2\rightarrow\mathbb R^2$ denotes the 2-D geometric transformation (to the reference frame) associated with the $i^{th}$ motion in the $j^{th}$ frame, which is represented by $L(i)_j$. To be specific, the $j^{th}$ row in $\tilde L(i)$ is $L(i)_j$ after certain permutation of its entries. The permutation results from applying the nonlinear transformation $\tau(i)_j$ to each nonzero pixel in $L(i)_j$ such that,
\begin{equation}
    \tau(i)_j(x,y)=(u,v),
    \end{equation}
    where $\tau(i)_j$ could be one of the five geometric transformations \cite{CVmodel}, i.e., translation, Euclidean, similarity, affine and homography, which are able to be represented by $2$, $3$, $4$, $6$ and $9$ free parameters, respectively. For example, affine transformation is defined as
    \begin{equation}
    \left[
      \begin{array}{c}
        u \\
        v \\
      \end{array}
    \right]
    =\left[
       \begin{array}{cc}
         \rho\cos\theta & \rho\sin\theta \\
         -\rho\sin\theta & \rho\cos\theta \\
       \end{array}
     \right]
     \left[
       \begin{array}{c}
         x \\
         y \\
       \end{array}
     \right]+
     \left[
       \begin{array}{c}
         t_x \\
         t_y \\
       \end{array}
     \right],
    \end{equation}
wherein $\theta$ is the rotation angle, $t_x$ and $t_y$ are the two translations and $\rho$ is the scaling ratio. It is worth to point out that $\tau(i)_j$ can be any other transformation beyond the geometric group. So SST can be applied to sparse structure in other applications if parametric form of $\tau(i)_j$ is known. We define the nonlinear operator $\circ$ as
    \begin{align}
    \notag \tilde L(i)_{j,u+(v-1)h}&=\left(L(i)_j\circ\tau(i)_j\right)_{u+(v-1)h}\\
    &=L(i)_{j,x+(y-1)h}.
    \end{align}
    Therefore, the flow tracking in SST aims at decomposing the sparse matrix $X$ ($S$ obtained in the background modeling) as
    \begin{equation}\label{E:SSTMODEL}
    \begin{array}{ll}
    &X=\sum\limits_{i=1}^k L(i)\circ\tau(i)+S+G,\\
    &{\rm rank}\left(L(i)\right)\leq r_i,{\rm card}(S)\leq s.
    \end{array}
    \end{equation}
    In SST, we iteratively invoke $k$ times of the following matrix decomposition to greedily construct the decomposition in (\ref{E:SSTMODEL}):
    \begin{equation}\label{E:SubMODEL}
    X=L\circ\tau+S+G,{\rm rank}\left(L\right)\leq r,{\rm card}(S)\leq s.
    \end{equation}
In each time of the matrix decomposition above, the data matrix $X$ is $S$ obtained by former decomposition. In order to save the computation and facilitate the parameter tuning, we cast the decomposition (\ref{E:SubMODEL}) into an optimization similar to (\ref{E:ls_app}),
    \begin{equation}\label{E:SSTOpt}
    \begin{array}{rl}
\min\limits_{L,\tau,S}&\|X-L\circ\tau-S\|_F^2+\lambda\|S\|_1\\
s.t.&{\rm rank}\left(L\right)\leq r,\\
\end{array}
\end{equation} 

Flow tracking in SST solves a sequence of optimization problem of type (\ref{E:SSTOpt}). Thus we firstly apply alternating minimization to (\ref{E:SSTOpt}). This results in iterative update of the solutions to the following three subproblems,
\begin{equation}\label{E:raw_td}
\left\{
  \begin{array}{ll}
    \tau^t=\arg\min\limits_\tau\|X-L^{t-1}\circ\tau-S^{t-1}\|_F^2; \\
    L^t=\arg\min\limits_{{\rm rank}\left(L\right)\leq r}\|X-L\circ\tau^t-S^{t-1}\|_F^2; \\
    S^t=\arg\min\limits_S\|X-L^t\circ\tau^t-S\|_F^2+\lambda\|S\|_1.
  \end{array}
\right.
\end{equation}

\subsubsection{Update of $\tau$}

The first subproblem aims at solving the following series of nonlinear equations of $\tau_j$,
\begin{equation}\label{E:tau}
L^{t-1}_j\circ\tau_j=X_j-S^{t-1}_j, j=1,\cdots,n.
\end{equation}
Albeit directly solving the above equation is difficult due to its strong nonlinearity, we can approximate the geometric transformation $L^{t-1}_j\circ\tau_j$ by using piece-wise linear transformations, where each piece corresponds to a small change of $\tau_j$ defined by $\Delta\tau_j$. Thus the solution of (\ref{E:tau}) can be approximated by accumulating a series of $\Delta\tau_j$. This can be viewed as an inner loop included in the update of $\tau$. Thus we have linear approximation
\begin{equation}\label{E:tauapp}
L^{t-1}_j\circ\left(\tau_j+\Delta\tau_j\right)\approx L^{t-1}_j\circ\tau_j+\Delta\tau_jJ_j,
\end{equation}
where $J_j$ is the Jacobian of $L^{t-1}_j\circ\tau_j$ with respect to the transformation parameters in $\tau_j$. Therefore, by substituting (\ref{E:tauapp}) into (\ref{E:tau}), $\Delta\tau_j$ in each linear piece can be solved as
\begin{equation}\label{E:tauequ}
\Delta\tau_j=\left(X_j-S^{t-1}_j-L^{t-1}_j\circ\tau_j\right)\left(J_j\right)^\dag.
\end{equation}
The update of $\tau_j$ starts from some initial $\tau_j$, and iteratively solves the overdetermined linear equation (\ref{E:tauequ}) with update $\tau_j:=\tau_j+\Delta\tau_j$ until the difference between the left hand side and the right hand side of (\ref{E:tau}) is sufficiently small. It is critical to emphasize that a well selected initial value of $\tau_j$ can significantly save computational time. Based on the between-frame affinity, we initialize $\tau_j$ by the transformation of its adjacent frame that is closer to the template frame $s$,
\begin{equation}
\tau_j:=\left\{
          \begin{array}{ll}
            \tau_{j+1}, & \hbox{$j<s$;} \\
            \tau_{j-1}, & \hbox{$j>s$.}
          \end{array}
        \right.
\end{equation}
Another important support set constraint, ${\rm supp}(L\circ\tau)\subseteq{\rm supp}(X)$, needs to be considered in calculating $L^{t-1}_j\circ\tau_j$ during the update of $\tau$. This constraint ensures that the object flows or segmented motions obtained by SST always belong to the sparse part achieved from the background modeling, and thus rules out the noise in background. Hence, suppose the complement set of ${\rm supp}(X_j)$ to be ${\rm supp}_c(X_j)$, each calculation of $L^{t-1}_j\circ\tau_j$ follows a screening such that,
\begin{equation}
\left(L^{t-1}_j\circ\tau_j\right)_{{\rm supp}_c(X_j)}=\overrightarrow{0}.
\end{equation}

\subsubsection{Update of $L$}

The second subproblem has the following global solution that can be updated by BRP based low-rank approximation (\ref{E:lr_app}) and its power scheme modification,
\begin{equation}
L^t=\sum\limits_{i=1}^r\lambda_iU_iV_i^T, {\rm svd}\left(\left(X-S^{t-1}\right)\circ\tau^{-1}\right)=U\Lambda V^T,
\end{equation}
wherein $\tau^{-1}$ denotes the inverse transformation towards $\tau$. The SVDs can be accelerated by BRP based low-rank approximation (\ref{E:ls_app}). Another acceleration trick is based on the fact that most columns of $\left(X-S^{t-1}\right)\circ\tau^{-1}$ are nearly all-zeros. This is because the object flow or motion after transformation occupies a very small area of the whole frame. Therefore, The update of $L^t$ can be reduced to low-rank approximation of a submatrix of $\left(X-S^{t-1}\right)\circ\tau^{-1}$ that only includes dense columns. Since the number of dense columns is far less than $p$, the update of $L^t$ can become much faster.

\subsubsection{Update of $S$}

The third subproblem has a global solution that can be obtained via soft-thresholding $\mathcal {P}_\lambda(\cdot)$ similar to the update of $S$ in GreBsmo,
\begin{equation}
S^t=\mathcal {P}_\lambda\left(X-L^t\circ\tau^t\right).
\end{equation}

\begin{algorithm}[H]\label{A:SST}\footnotesize
\SetAlgoLined
\KwIn{$X$, $r_i,\lambda_i(i=1,\cdots,n)$, $k$}
\KwOut{$L_i(i=1,\cdots,n)$, $S$}
\For{$i\leftarrow 1$ \KwTo $k$}{
\textbf{Initialize:} $s=\arg\max\limits_i {\rm card}\left(X_i\right)$\;
$L=\left[X_s;\cdots;X_s\right]$, $S=\textbf{0}$, $\tau=\overrightarrow{0}$\;
\While{not converge}{
\For{$j\leftarrow s-1$ \KwTo $1$}{
$\tau_j:=\tau_{j+1}$\;
\While{not converge}{
$\tilde L^{t-1}_j=L^{t-1}_j\circ\tau_j, \tilde L^{t-1}_{j,{\rm supp}_c(X_j)}=\overrightarrow{0}$\;
$\tau_j:=\tau_j+\left(X_j-S^{t-1}_j-\tilde L^{t-1}_j\right)\left(J_j\right)^\dag$\;
}
}
\For{$j\leftarrow s+1$ \KwTo $n$}{
$\tau_j:=\tau_{j-1}$\;
\While{not converge}{
$\tilde L^{t-1}_j=L^{t-1}_j\circ\tau_j, \tilde L^{t-1}_{j,{\rm supp}_c(X_j)}=\overrightarrow{0}$\;
$\tau_j:=\tau_j+\left(X_j-S^{t-1}_j-\tilde L^{t-1}_j\right)\left(J_j\right)^\dag$\;
}
}
$\tau^t=\tau$\;
$L^t={\rm BRP}\left(\left(X-S^{t-1}\right)\circ\tau^{-1}\right)$\;
$S^t=\mathcal {P}_\lambda\left(X-L^t\circ\tau^t\right), S^t_{j,{\rm supp}_c(X_j)}=\overrightarrow{0}$\;
}
$X:=S^t, L(i):=L^t, \tau(i)=\tau^t$\;
}
\caption{Shifted Subspace Tracking (SST)}
\end{algorithm}\normalsize

A support set constraint ${\rm supp}(S)\subseteq{\rm supp}(X)$ should be considered in the update of $S$ as well. Hence the above update follows a postprocessing,
\begin{equation}
S^t_{j,{\rm supp}_c(X_j)}=\overrightarrow{0}, j=1,\cdots,n.
\end{equation}

Note the transformation computation $\circ$ in the update can be accelerated by leveraging the sparsity of the motions. Specifically, the sparsity allows SST to only compute the transformed positions of the nonzero pixels. We summarize the SST algorithm in Algorithm \ref{A:SST}.

\subsection{Multi-label Subspace Ensemble}

MSE provides a novel insight into the multi-label learning (ML) problem, which aims at predicting multiple labels of a data sample. Most previous ML methods \cite{CC, HBR, MDDM, SharedSubspace, RAkEL, Stacking} focus on training effective classifiers that establishes a mapping from feature space to label space, and take the label correlation into account in the training process. Because it has been longly believed that label correlation is useful for improving prediction performance. However, in these methods, both the label space and the model complexity will grow rapidly when increasing the number of labels and simultaneously modeling their joint correlations. This usually makes the available training samples insufficient for learning a joint prediction model. 

MSE eliminates this problem by jointly learning inverse mappings that map each label to the feature space as a subspace, and formulating the prediction as finding the group sparse representation \cite{GroupLasso} of a given sample on the ensemble of subspaces. In the training stage, the training data matrix $X$ is decomposed as the sum of several low-rank matrices and a sparse residual via a randomized optimization. Each low-rank part defines a subspace mapped by a label, and its rows are nonzero only when the corresponding samples are annotated by the label. The sparse part captures the rest contents in the features that cannot be explained by the labels. 

%

\subsubsection{MSE training: randomized decomposition}

The training stage of MSE approximately decomposes the training data matrix $X\in\mathbb R^{n\times p}$ into $X=\sum_{i=1}^kL^i+S$. For the matrix $L^i$, the rows corresponding to the samples with label $i$ are nonzero, while the other rows are all-zero vectors. The nonzero rows denote the components explained by label $i$ in the feature space. We use $\Omega_i$ to denote the index set of samples with label $i$ in the matrix $X$ and $L^i$, and then the matrix composed of the nonzero rows in $L^i$ is represented by $L^i_{\Omega_i}$. In the decomposition, the rank of $L^i_{\Omega_i}$ is upper bounded, which indicates that all the components explained by label $i$ nearly lies in a linear subspace. The matrix $S$ is the residual of the samples that cannot be explained by the given labels. In the decomposition, the cardinality of $S$ is upper bounded, which makes $S$ sparse.

If the label matrix of $X$ is $Y\in\{0,1\}^{n\times k}$, the rank of $L^i_{\Omega_i}$ is upper bounded by $r^i$ and the cardinality of $S$ is upper bounded by $K$, the decomposition can be written as solving the following constrained minimization problem:
\begin{equation}\label{E:ms}
\begin{array}{rl}
\min\limits_{L^i,S}&\left\|X-\sum_{i=1}^kL^i-S\right\|_F^2\\
s.t.&{\rm rank}\left(L^i_{\Omega_i}\right)\leq r^i,L^i_{\overline{\Omega}_i}=\textbf{0},\forall i=1,\dots,k\\
&{\rm card}\left(S\right)\leq K.
\end{array}
\end{equation}
Therefore, each training sample in $X$ is decomposed as the sum of several components, which respectively correspond to multiple labels that the sample belongs to. MSE separates these components from the original sample by building the mapping from the labels to the feature space. For label $i$, we obtain its mapping in the feature space as the row space of $L^i_{\Omega_i}$.

Although the rank constraint to $L^i_{\Omega_i}$ and cardinality constraint to $S$ are not convex, the optimization in (\ref{E:ms}) can be solved by alternating minimization that decomposes it as the following $k+1$ subproblems, each of which has the global solution:
\begin{equation}\label{E:mssub}
\left\{
  \begin{array}{ll}
    L^i_{\Omega_i}=\arg\min\limits_{{\rm rank}\left(L^i_{\Omega_i}\right)\leq r^i}\left\|X-\sum\limits_{j=1,j\neq i}^kL^j-S-L^i\right\|_F^2, \\
    ~~~~~~~~\forall i=1,\dots,k.\\
    S=\arg\min\limits_{{\rm card}\left(S\right)\leq K}\left\|X-\sum\limits_{j=1}^kL^j-S\right\|_F^2.
  \end{array}
\right.
\end{equation}

The solutions of $L^i_{\Omega_i}$ and $S$ in the above subproblems can be obtained via hard thresholding of singular values and the matrix entries, respectively. Note that both SVD and matrix entry-wise hard thresholding have global solutions. In particular, $L^i_{\Omega_i}$ is built from the first $r^i$ largest singular values and the corresponding singular vectors of $\left(X-\sum_{j=1,j\neq i}^kL^j-S\right)_{\Omega_i}$, while $S$ is built from the $K$ entries with the largest absolute value in $X-\sum_{j=1}^kL^j$, i.e.,
\begin{equation}\label{E:mssolution}
\left\{
  \begin{array}{ll}
    L^i_{\Omega_i}=\sum\limits_{q=1}^{r^i}\lambda_qU_qV_q^T, i=1,\dots,k,\\
    {\rm svd}\left[\left(X-\sum_{j=1,j\neq i}^kL^j-S\right)_{\Omega_i}\right]=U\Lambda V^T; \\
    S=\mathcal {P}_{\Phi}\left(X-\sum\limits_{j=1}^kL^j\right), \Phi:\left|\left(X-\sum\limits_{j=1}^kL^j\right)_{{r,s}\in{\Phi}}\right|\neq0 \\ {\rm~and~} \geq \left|\left(X-\sum\limits_{j=1}^kL^j\right)_{{r,s}\in{\overline{\Phi}}}\right|, |\Phi|\leq K.
  \end{array}
\right.
\end{equation}
The projection $S=\mathcal {P}_{\Phi}(R)$ represents that the matrix $S$ has the same entries as $R$ on the index set $\Phi$, while the other entries are all zeros.

The decomposition is then obtained by iteratively solving these $k+1$ subproblems in (\ref{E:mssub}) according to (\ref{E:mssolution}). In this problem, we initialize $L^i_{\Omega_i}$ and $S$ as
\begin{equation}\label{E:msinitial}
\left\{
  \begin{array}{ll}
  L^i_{\Omega_i}:=Z_{\Omega_i},i=1,\dots,k,\\
  Z=D^{-1}X,D={\rm diag}\left(Y\textbf{1}\right);\\
  S:=\textbf{0}.
  \end{array}
\right.
\end{equation}
In each subproblem, only one variable is optimized with the other variables fixed. Similar to GoDec, BRP based acceleration strategy can be applied to the above model and produces the practical training algorithm in Algorithm \ref{A:MSEtraining}.

In the training, the label correlations is naturally preserved in the subspace ensemble, because all the subspaces are jointly learned. Since only $k$ subspaces are learned in the training stage, MSE explores label correlations without increasing the model complexity.

\begin{algorithm}[H]\label{A:MSEtraining}\footnotesize
\SetAlgoLined
\KwIn{$X$, $\Omega_i$, $r^i,i=1,\dots,k$, $K$, $\epsilon$}
\KwOut{$C^i,i=1,\dots,k$}
Initialize $L^i$ and $S$ according to (\ref{E:msinitial}), $t:=0$\;
\While{$\left\|X-\sum_{j=1}^kL^j-S\right\|_F^2>\epsilon$}{
$t:=t+1$\;
\For{$i\leftarrow 1$ \KwTo $k$}{
$N:=\left(X-\sum_{j=1,j\neq i}^kL^j-S\right)_{\Omega_i}$\;
Generate standard Gaussian matrix $A_1\in\mathbb R^{p\times{r^i}}$\;
$Y_1:=NA_1$, $A_2:=Y_1$\;
$Y_2:=N^TY_1$, $Y_1:=NY_2$\;
$L^i_{\Omega_i}:=Y_1\left(A_2^TY_1\right)^{-1}Y_2^T, L^i_{\overline{\Omega}_i}:=\textbf{0}$\;
}
$N:=\left|X-\sum_{j=1}^kL^j\right|$\;
$S:=\mathcal {P}_{\Phi}\left(N\right)$, $\Phi$ is the index set of the first $K$ largest entries of $\left|N\right|$\;
}
QR decomposition $\left(L^i_{\Omega_i}\right)^T=Q^iR^i$ for $i=1,\dots,k$, $C^i:=\left(Q^i\right)^T$\;
\caption{MSE Training}
\end{algorithm}\normalsize

\subsubsection{MSE prediction: group sparsity}


In the prediction stage of MSE, we use group \emph{lasso} \cite{GroupLasso}\cite{SLEP} to estimate the group sparse representation $\beta\in\mathbb R^{\sum {r^i}}$ of a test sample $x\in\mathbb R^p$ on the subspace ensemble $C=[C^1;\dots;C^k]$, wherein the $k$ groups are defined as index sets of the coefficients corresponding to $C^1,\dots,C^k$. Since group \emph{lasso} selects nonzero coefficients group-wisely, nonzero coefficients in the group sparse representation will concentrate on the groups corresponding to the labels that the sample belongs to.

According to the above analysis, we solve the following group \emph{lasso} problem in the prediction stage of MSE
\begin{equation}\label{E:mspredict}
\min\limits_\beta \frac{1}{2}\left\|x-\beta C\right\|_F^2+\lambda\sum\limits_{i=1}^k\left\|\beta_{G_i}\right\|_2,\\
\end{equation}
where the index set $G_i$ includes all the integers between $1+\sum_{j=1}^{i-1}r^j$ and $\sum_{j=1}^{i}r^j$ (including these two).

To obtain the final prediction of the label vector $y\in\{0,1\}^k$ for a test sample $x$, we use a simple thresholding of the magnitude sum of coefficients in each group to test which groups that the sparse coefficients in $\beta$ concentrate on
\begin{equation}\label{E:thresh}
y_\Psi=\textbf{1},y_{\overline\Psi}=\textbf{0},\Psi=\left\{i:\left\|\beta_{G_i}\right\|_1\geq\delta\right\}.
\end{equation}
Although $y$ can also be obtained via selecting the groups with nonzero coefficients when $\lambda$ in (\ref{E:mspredict}) is chosen properly, we set the threshold $\delta$ as a small positive value to guarantee the robustness to $\lambda$.


\subsection{Linear Functional GoDec for Learning Recommendation System}

Although low-rank matrix completion provides an effective and simple mathematical model predicting a user's rating to an item from her/his ratings to other items and the ratings of other users by exploring the user relationships, a primary problem of this model is that adding a new item or a new user to the model requires an new optimization of the whole low-rank rating matrix, which is not practical due to its expensive time cost. Moreover, although the attributes of users are always missing in real recommendation systems, features of the items have been proved to be helpful side information that is much easier to obtain. But previous matrix completion methods and GoDec cannot leverage this information in their models. Furthermore, robust rating prediction should allow advertising effects in known ratings. 

In this part, we propose a variant of GoDec called ``linear functional GoDec (LinGoDec)''. It formulates the collaborative filtering problem as supervised learning, and avoids time-consuming completion of the whole matrix when only a new item's scores (a new row) are needed to be predicted. In particular, LinGoDec decomposes rating matrix $X$ whose rows index the users, columns index the items, and entries denote the scores of items given by different users. Given the features of some items, which are usually available, and the ratings of these items scored by all users, LinGoDec learns a scoring function for each user so that efficient prediction of ratings can be made item-wisely. It studies the case when the scoring functions of different users are linear and related to each other. In the mode, it replaces the low-rank part $L$ of GoDec with $WZ^T$, where $W$ represents the linear related functions and the rows of $Z$ are items represented by features. The sparse part $S$ is able to capture the advertising effects or anomaly of users' ratings on specific items, which cannot be represented by the low-rank scoring functions. In the algorithm of LinGoDec, the update of low-rank $W$ is accomplished by invoking an elegant closed-form solution for least square rank minimization \cite{ALS}, which could be accelerated by BRP.

LinGoDec aims at solving the following optimization,
\begin{equation}\label{equ:LinGoDec}
\begin{array}{ll}
&\min_{W,S}\|X-WZ^T-S\|_F^2+\lambda\|{\rm vec}(S)\|_1\\
&{\rm s.t.}~~rank(W)\leq r.
\end{array}
\end{equation}

We constrain $W$ to be low-rank so that the functions of different users share the same small set of basis functions. In addition, we apply $\ell_1$ regularization to the entries of $S$ so that the advertising effects in training ratings can be captured and ruled out from the learning of $W$. By applying alternating minimization to (\ref{equ:LinGoDec}), we have
\begin{equation}\label{equ:NLinGoDec}
\left\{
  \begin{array}{ll}
    W_k=\arg\min_W \left\|X-WZ^T\right\|_F^2~~s.t.~~{\rm rank}(W)\leq r, \\
    S_k=\mathcal S_\lambda\left(X-W_kZ^T\right),
  \end{array}
\right.
\end{equation}
The update of $W_k$ in above procedures equals to solve a least squares rank minimization, which has been discovered owning closed-form solution that can be obtained by truncated SVD [] when $X$ is singular (the most common case in our problem). By applying bilateral random projection based acceleration to the truncated SVD, we immediately achieve the final fast algorithm for LinGoDec. LinGoDec has a similar model as rank-regularized multi-task learning, but the major difference is that the sparse matrix in LinGoDec is a component of the data matrix rather than the linear functions $W$. 

\section{Experiments}

This section evaluates both the effectiveness and the efficiency of all the algorithms proposed in this paper, and compares them with state-of-the-art rivals. We will show experimental results of GoDec and GreBsmo on both surveillance video sequences for background modeling and synthetic data. Then we will apply SST, MSE and LinGoDec to the problems of motion segmentation, multi-label learning and collaborative filtering. We run all the experiments in MatLab on a server with dual quad-core 3.33 GHz Intel Xeon processors and 32 GB RAM. The relative error $\|X-\hat X\|_F^2/\|X\|_F^2$ is used to evaluate the effectiveness, wherein $X$ is the original matrix and $\hat X$ is an estimate/approximation.

\begin{table}[H]\footnotesize
\renewcommand{\arraystretch}{1}
\caption{Relative error and time cost of RPCA and GoDec in low-rank+sparse decomposition tasks. The results separated by ``$/$'' are RPCA and GoDec, respectively.}
\begin{center}\vspace{-1mm}
\begin{tabular}{ccc|cccc}
\hline
${\rm size}(X)$ & ${\rm rank}(L)$ & ${\rm card}(S)$ & ${\rm rel.error}(X)$ & ${\rm rel.error}(L)$ & ${\rm rel.error}(S)$ & ${\rm time}$\\
(square)&$(1)$&$(10^4)$&$(10^{-8})$&$(10^{-8})$&$(10^{-6})$&(seconds)\\
\hline\hline
500   & 25  & $1.25$ & $3.70/1.80$ & $1.50/1.20$ & $2.00/0.95$ & $6.07/2.83$\\
1000  & 50  & $5.00$ & $4.98/4.56$ & $1.82/1.85$ & $5.16/4.90$ & $20.96/12.71$\\
2000  & 100 & $20.0$ & $8.80/1.13$ & $3.10/1.10$ & $1.81/1.24$ & $101.74/74.16$\\
3000  & 250 & $45.0$ & $6.29/4.98$ & $5.09/5.05$ & $33.9/55.3$ & $562.09/266.11$\\
5000  & 400 & $125$  & $63.1/24.4$ & $30.2/29.3$ & $54.2/18.8$ & $2495.31/840.39$\\
10000 & 500 & $600$  & $6.18/3.04$ & $2.27/2.88$ & $58.3/36.6$ & $9560.74/3030.15$\\
\hline
\end{tabular}\label{table:RPCAvsGo}
\end{center}\vspace{-1mm}
\end{table}

\subsection{GoDec on Synthetic Data}

We compare the relative errors and time costs of Robust PCA and GoDec on square matrices with different sizes, different ranks of low-rank components and different cardinality of sparse components. For a matrix $X=L+S+G$, its low-rank component is built as $L=AB$, wherein both $A$ and $B$ are $n\times r$ standard Gaussian matrices. Its sparse part is built as $S=\mathcal P_\Omega(D)$, wherein $D$ is a standard Gaussian matrix and $\Omega$ is an entry set of size $k$ drawn uniformly at random. Its noise part is built as $G=10^{-3}\cdot F$, wherein $F$ is a standard Gaussian matrix. In our experiments, we compare RPCA \footnote{http://watt.csl.illinois.edu/\~perceive/matrix-rank} (\texttt{inexact\_alm\_rpca}) with GoDec (Algorithm \ref{alg:GoDec} with $q=2$). Since both algorithms adopt the relative error of $X$ as the stopping criterion, we use the same tolerance $\epsilon=10^{-7}$. Table \ref{table:RPCAvsGo} shows the results and indicates that both algorithms are successful in recovering the correct ``low-rank+sparse'' decompositions with relative error less than $10^{-6}$. GoDec usually produces less relative error with much less CPU seconds than RPCA. The improvement of accuracy is due to that the model of GoDec in (\ref{E:GoDecMODEL}) is more general than that of RPCA by considering the noise part. The improvement of speed is due to that BRP based low-rank approximation significantly saves the computation of each iteration round.

\subsection{GoDec for Background Modeling}

\begin{figure*}[ht]
\begin{center}\vspace{-1.7mm}
 \includegraphics[width=0.84\linewidth]{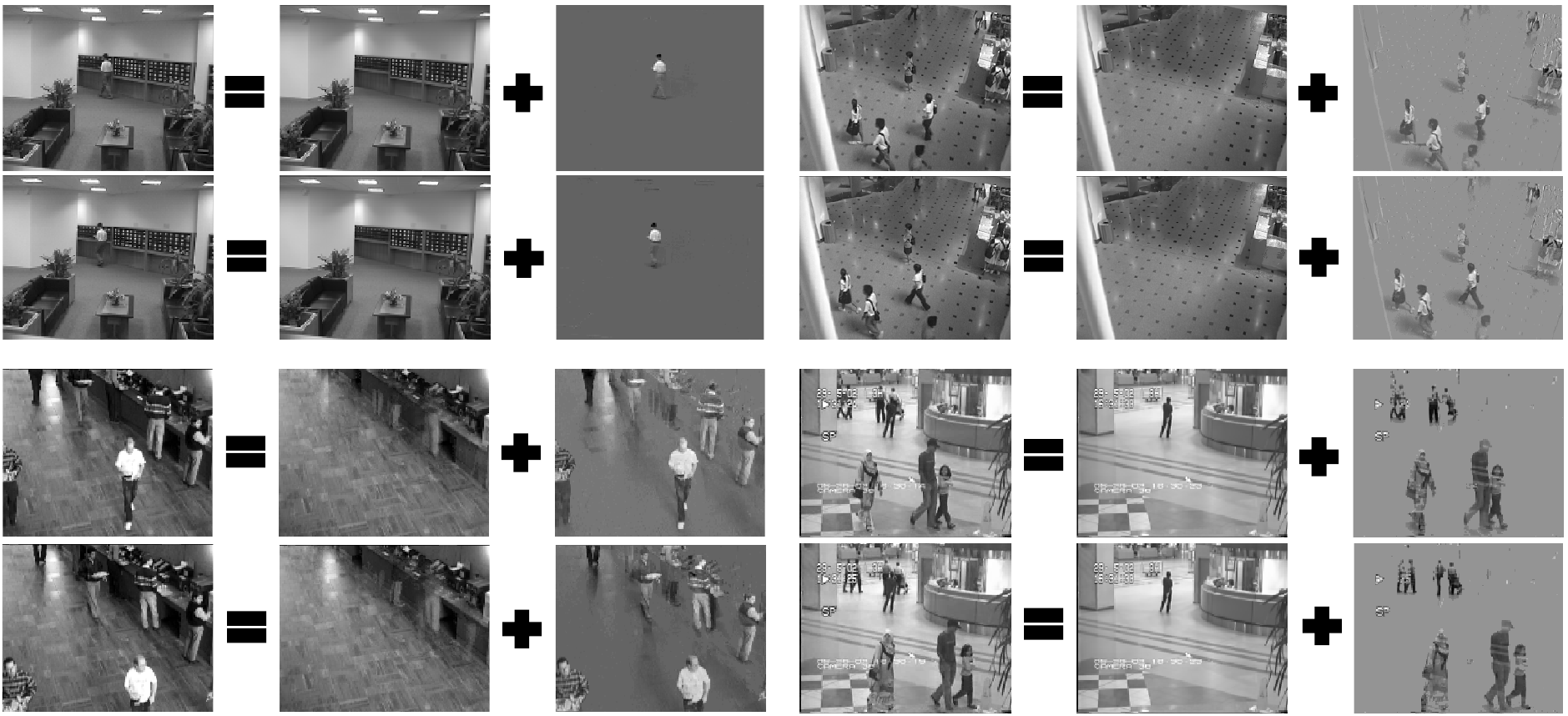}
\end{center}\vspace{-1.5mm}
   \caption{Background modeling results of four $200$-frame surveillance video sequences in $X=L+S$ mode. Top left: lobby in an office building (resolution $128\times 160$, learning time $39.75$ seconds). Top right: shopping center (resolution $256\times 320$, learning time $203.72$ seconds). Bottom left: Restaurant (resolution $120\times 160$, learning time $36.84$ seconds). Bottom right: Hall of a business building (resolution $144\times 176$, learning time $47.38$ seconds).}
\label{fig:background}
\end{figure*}\vspace{-2mm}

Background modeling \cite{BackgroundS} is a challenging task to reveal the correlation between video frames, model background variations and foreground moving objects. A video sequence satisfies the low-rank+sparse structure, because backgrounds of all the frames are related, while the variation and the moving objects are sparse and independent. We apply GoDec (Algorithm \ref{alg:GoDec} with $q=2$) to four surveillance videos \footnote{http://perception.i2r.a-star.edu.sg/bk\_model/bk\_index.html}, respectively. The matrix $X$ is composed of the first $200$ frames of each video. For example, the second video is composed of $200$ frames with the resolution $256\times 320$, we convert each frame as a vector and thus the matrix $X$ is of size $81920\times 200$. We show the decomposition result of one frame in each video sequence in Figure \ref{fig:background}. The background and moving objects are precisely separated (the person in $L$ of the fourth sequence does not move throughout the video) without losing details. The results of the first sequence and the fourth sequence are comparable with those shown in \cite{RobustPCA}. However, compared with RPCA ($36$ minutes for the first sequence and $43$ minutes for the fourth sequence) \cite{RobustPCA}, GoDec requires around $50$ seconds for each of both. Therefore, GoDec makes large-scale applications available.

\subsection{GoDec for Shadow/Light removal}

\begin{figure*}[h!]
\begin{center}\vspace{-1.7mm}
 \includegraphics[width=1\linewidth]{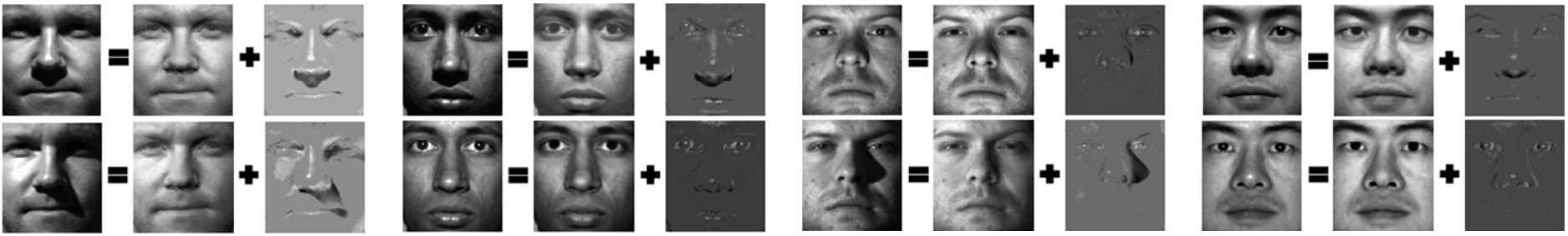}
\end{center}\vspace{-1.5mm}
   \caption{Shadow/light removal of face images from four individuals in Yale B database in $X=L+S$ mode. Each individual has $64$ images with resolution $192\times 168$ and needs $24$ seconds learning time.}
\label{fig:face}
\end{figure*}

Shadow and light in training images always pull down the quality of learning in computer vision applications. GoDec can remove the shadow/light noises by assuming that they are sparse and the rest parts of the images are low-rank. We apply GoDec (Algorithm \ref{alg:GoDec} with $q=2$) to face images of four individuals in the Yale B database \footnote{http://cvc.yale.edu/projects/yalefacesB/yalefacesB.html}. Each individual has $64$ images with resolution $192\times 168$ captured under different illuminations. Thus the matrix $X$ for each individual is of size $32760\times 64$. We show the GoDec of eight example images ($2$ per individual) in Figure \ref{fig:face}. The real face of each individual are remained in the low rank component, while the shadow/light noises are successfully removed from the real face images and stored in the sparse component. The learning time of GoDec for each individual is less than $30$ seconds, which encourages for large-scale applications, while RPCA requies around $685$ seconds.

\vspace{-2mm}

\subsection{GreBsmo on Synthetic Data}

\begin{figure*}[htb]\label{fig:GreBsmoexp}
\begin{center}\vspace{-3mm}
\subfigure{
 \includegraphics[width=0.35\linewidth]{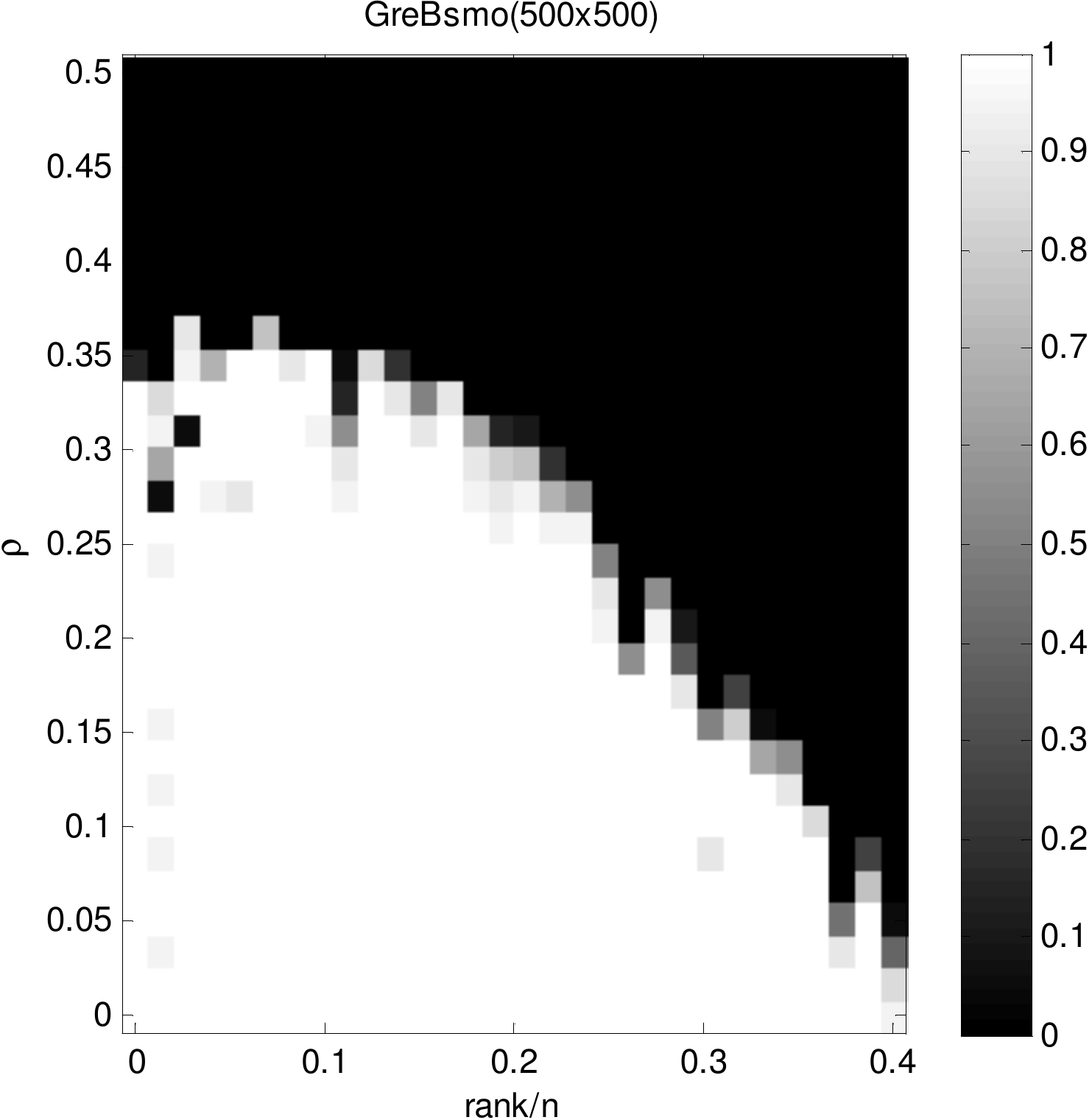}
}
\subfigure{
 \includegraphics[width=0.6\linewidth]{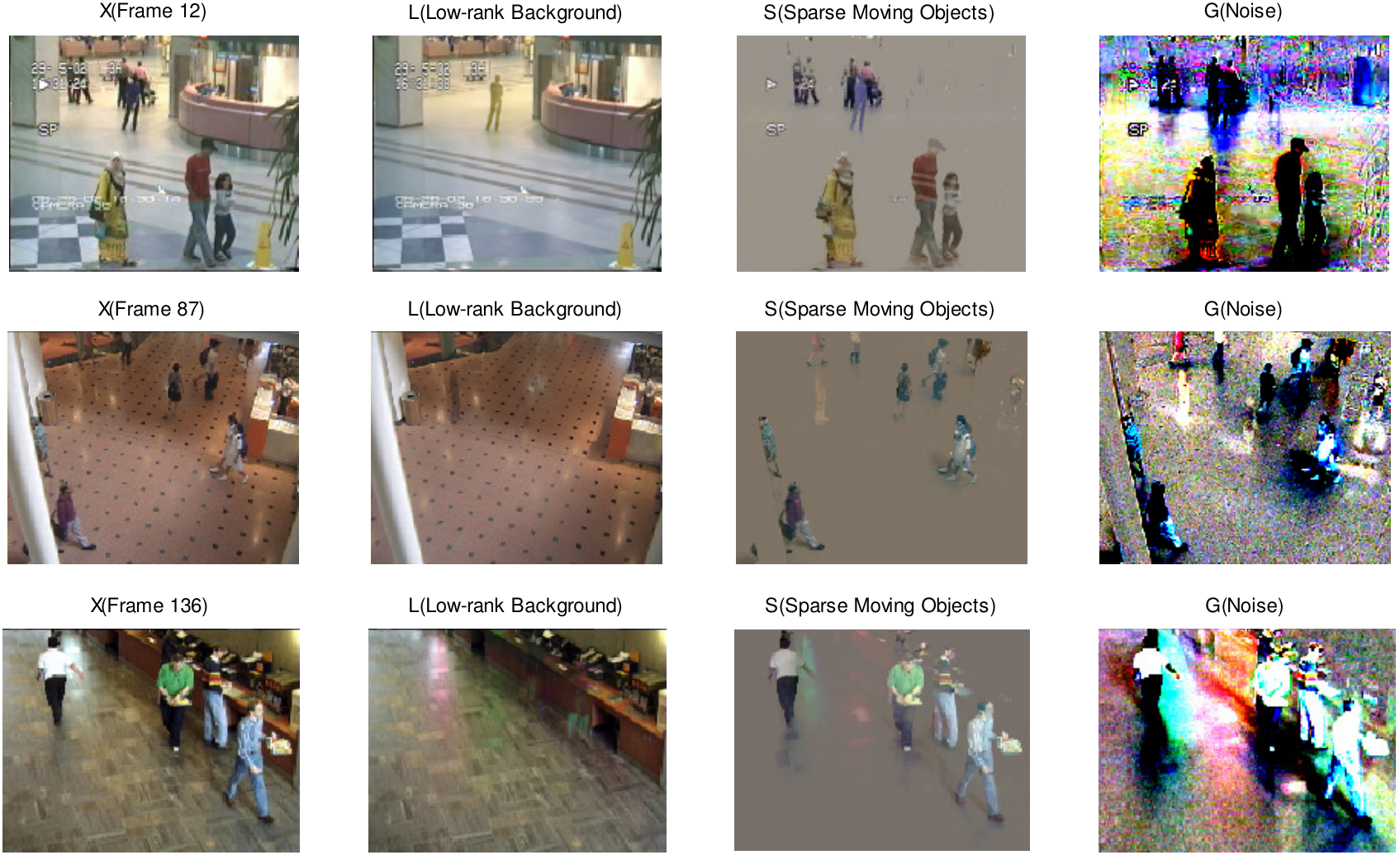}
}
\end{center}\vspace{-5mm}
\caption{Phase diagram for GreBsmo (left) on $500\times 500$ matrices. Low-rank component is generated as $L=UV$, where entries of $U$ and $V$ are sampled from $\mathcal N(0,1/n)$. Entries of sparse component $S$ are sampled as $1$ or $-1$ with probability $\rho/2$ and $0$ with probability $1-\rho$. On the $30\times 30$ grid of sparsity-rank/n plane, $20$ trials are performed for each $(\rho,r)$ pair. $L$ is said to be successfully recovered if its rel. err.$\leq 10^{-2}$. The phase diagram shows the successful recovery rate for each $(\rho,r)$ pair.
Background modeling of GreBsmo (right) on three video sequences, top row: Hall, $144\times 176$ pixels, $500$ frames; middle row: ShoppingMall, $256\times 320$ pixels, $253$ frames; bottom row: Boostrap, $120\times 160$ pixels, $500$ frames.}
\end{figure*}

We report the phase diagram of GreBsmo in Figure \ref{fig:GreBsmoexp} from results on randomly generated matrix that is the sum of a low-rank part and a sparse part. The low-rank part is generated as the product of two Gaussian matrices and the sparse part has a Bernoulli model generated support set on which $\pm 1$ values are randomly assigned. The phase transition phenomenon is in consistency with existing low-rank and sparse decomposition algorithms. It also shows that GreBsmo is able to gain accurate separation of $L$ even if its rank is close to $0.4n$, given the sparse part has an adequately sparse support set. This is competitive to published result \cite{RobustPCA}. Interestingly, the phase transition curve has a regular shape and implies a theoretical analysis to its behavior is highly possible in future studies.


\begin{table}[H]
\renewcommand{\arraystretch}{1}
\caption{Comparison of time costs in CPU seconds of PCP, GoDec and GreBsmo in low-rank and sparse matrix decomposition task on background modeling datasets.}
\begin{center}
\begin{tabular}{l|ccc}
\hline
&PCP &GoDec &GreBsmo\\
\hline
Hall &$87s$ &$56s$ &$1.13s$\\
ShoppingMall &$351s$ &$266s$ &$3.29s$\\
Bootstrap &$71s$ &$49s$ &$0.98s$\\
\hline
\end{tabular}\label{table:RPCAtime}
\end{center}\vspace{-3mm}
\end{table}
\normalsize

\subsection{GreBsmo for Background Modeling}

For real data, three robust PCA algorithms, i.e., inexact augmented Lagrangian multiplier method for PCP, GoDec and GreBsmo are applied to separate the low-rank background and sparse moving objects in 3 video sequences from the same dataset used in GoDec experiment above. We show the robust PCA decomposition results of one frame for each video sequence obtained by GreBsmo in the left plot of Figure \ref{fig:GreBsmoexp}. The time costs for all the three methods are listed in Table \ref{table:RPCAtime}. It shows GreBsmo considerably speed up the decomposition and performs $30$-$100$ times faster than most existing algorithms.


\subsection{SST for Motion Segmentation}

\begin{figure*}[htb]\label{fig:hallshop}
\begin{center}\vspace{-3mm}
\subfigure{
 \includegraphics[width=0.45\linewidth]{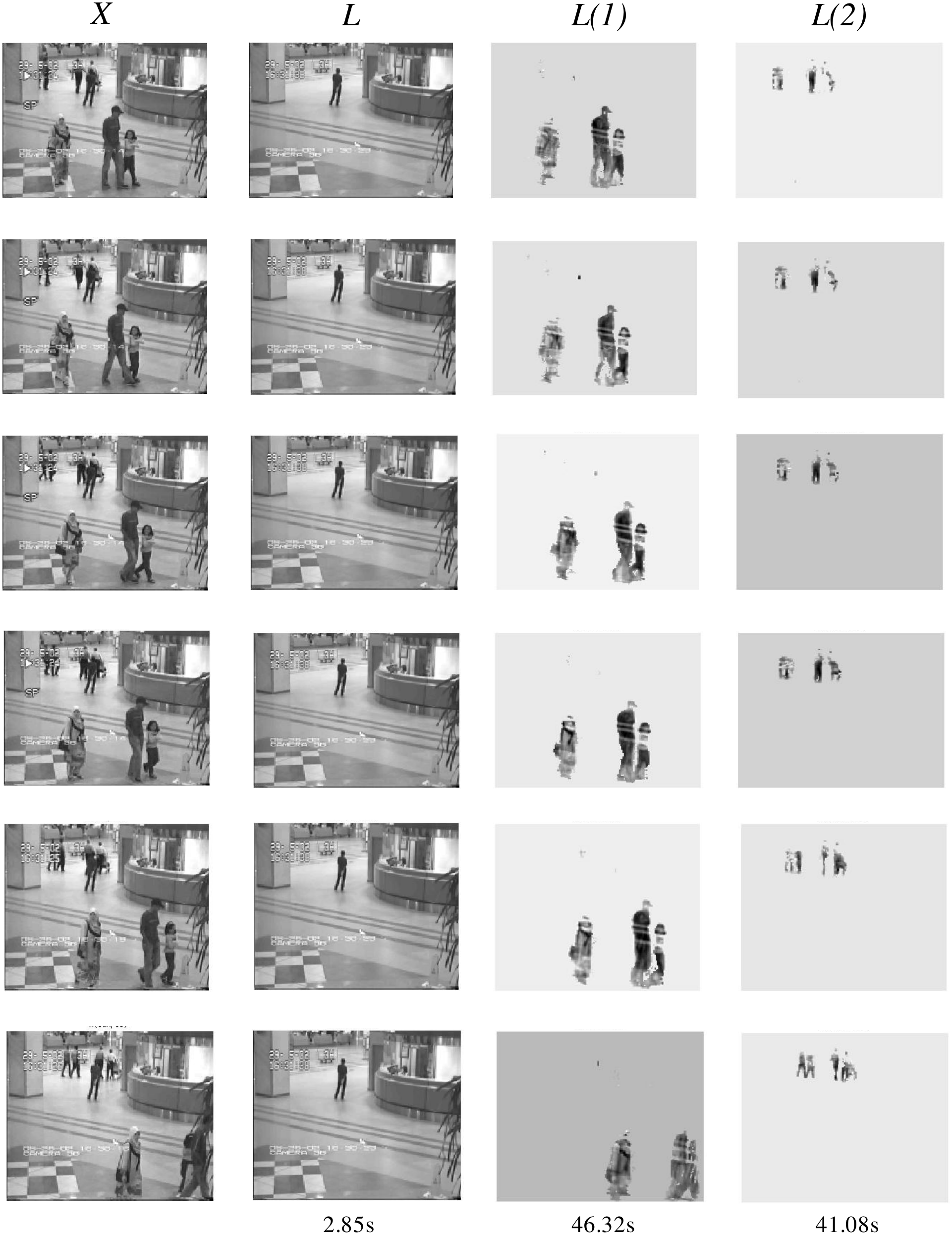}
}
\subfigure{
 \includegraphics[width=0.45\linewidth]{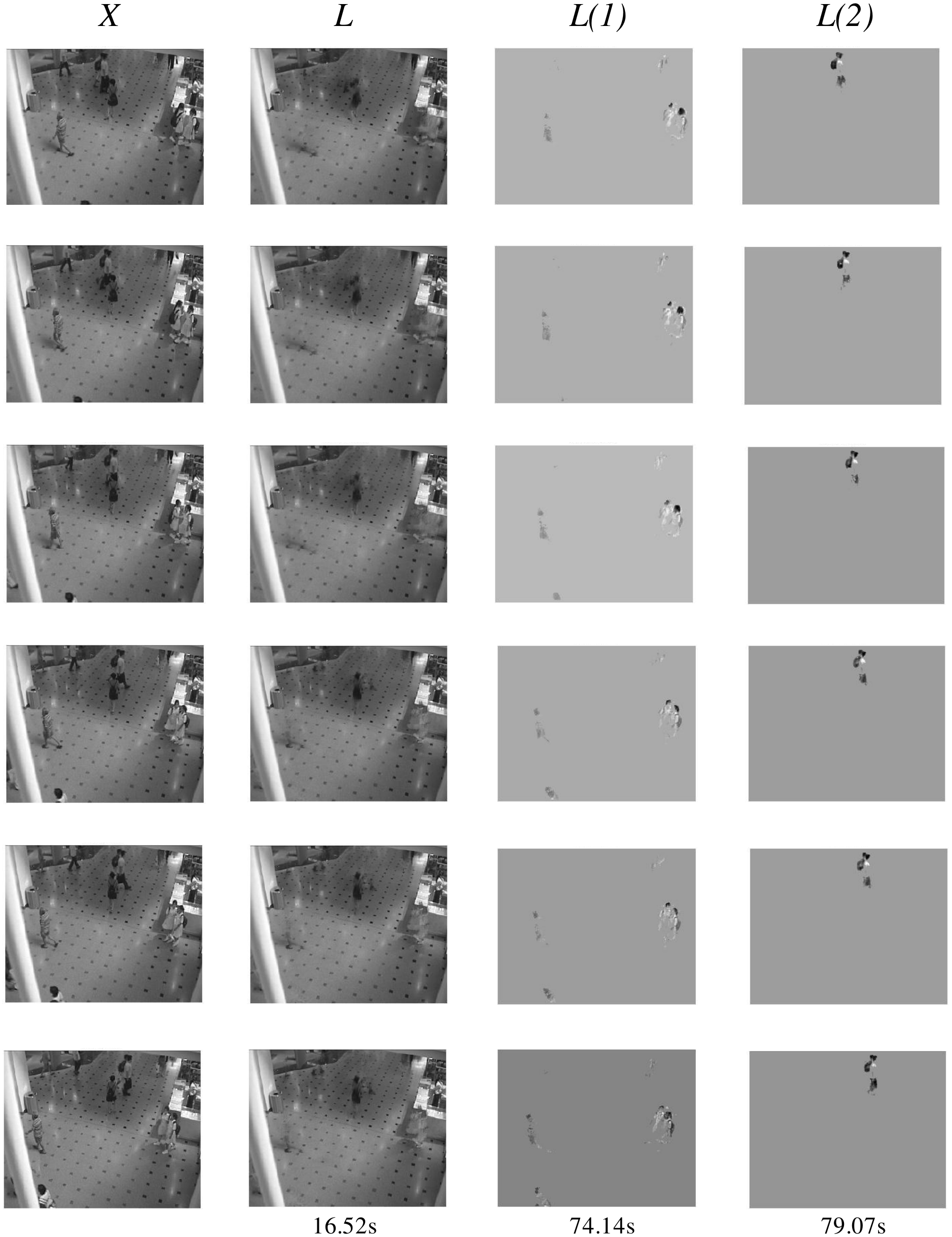}
}
\end{center}\vspace{-5mm}
\caption{Background modeling and object flow tracking results of a $50$-frame surveillance video sequence from Hall dataset with resolution $144\times 176$ (left), and Shoppingmall dataset with resolution $256\times 320$ (right).}
\end{figure*}

We evaluate SST by using it to track object flows in four surveillance video sequences from the same dataset. In these experiments, the type of geometric transformation $\tau$ is simply selected as translation. The detection, tracking and segmentation results as well as associated time costs are shown in Figure \ref{fig:hallshop}.

%

The results show SST can successfully recover both the low-rank patterns and the associated geometric transformations for motions of multiple object flows from the sparse component achieved by GoDec. The detection, tracking and segmentation are seamlessly unified in a matrix factorization framework and achieved with high accuracy. Moreover, it also verifies that SST performs significantly robust on complicated motions in complex scenes. This is attributed to their distinguishing shifted low-rank patterns, because different object flows can hardly share a subspace after the same geometric transformation. Since SST show stable and appealing performance in motion detection, tracking and segmentation for either crowd or individual, it provides a more semantic and intelligent analysis to the video content than existing methods.

\subsection{MSE for Multi-label Learning}

We evaluate MSE on 13 benchmark datasets from different domains and of different scales, including Corel5k (image), Scene (image), Mediamill (video), Enron (text), Genbase (genomics), Medical (text), Emotions (music), Slashdot (text) and $5$ sub datasets selected in Yahoo dataset (web data). These datasets were obtained from Mulan's website \footnote{\texttt{http://mulan.sourceforge.net/datasets.html}} and MEKA's website \footnote{\texttt{http://meka.sourceforge.net/}}. They were collected from different practical problems.

We compare MSE with BR \cite{MLreview2}, ML-KNN \cite{MLKNN} and MDDM \cite{MDDM} on four evaluation metrics for evaluating the effectiveness, as well as the CPU seconds for evaluating the efficiency. In multi-label prediction, four metrics, which are precision, recall, F1 score and accuracy, are used to measure the prediction performance. The detailed definitions of these metrics are given in Section 7.1.1 of \cite{MLreview1}. A fair evaluation of prediction performance should include integrative consideration of all the four metrics, whose importances can be roughly given by $F1, Acc>\{Prec, Rec\}$.

We show the prediction performance and time cost in CPU seconds of BR, ML-KNN, MDDM and MSE in Table \ref{Table:exp} and Table \ref{Table:yahoo}. In BR, we use the MatLab interface of LIBSVM 3.0 \footnote{\texttt{http://www.csie.ntu.edu.tw/\~cjlin/libsvm/}} to train the classic linear SVM classifiers for each label. The parameter $C\in\left\{10^{-3},10^{-2},0.1,1,10,10^2,10^3\right\}$ with the best performance on the training set was used. In ML-KNN, the number of neighbors was $30$ for all the datasets.

In MDDM, the regularization parameter for uncorrelated subspace dimensionality reduction was selected as $0.12$ and the dimension of the subspace was set as $20\%$ of the dimension of the original data. In MSE, we selected $r^i$ as an integer in $\left[1,6\right]$, $K\in\left[10^{-6},10^{-3}\right]$, $\lambda\in\left[0.2,0.45\right]$ and $\delta\in\left[10^{-4},10^{-2}\right]$. We roughly selected $4$ groups of parameters in the ranges for each dataset and chose the one with the best performance on the training data. Group \emph{lasso} in MSE is solved by SLEP \cite{SLEP} in our experiments.

The experimental results show that MSE is competitive on both speed and prediction performance, because it explores label correlations and structure without increasing the problem size. In addition, the bilateral random projections further accelerate the computation. In particular, its training time increases much more slowly than other methods, so it is more efficient when applied to large scale datasets such as Mediamill, Arts and Education. MDDM is faster than MSE on a few datasets because MDDM invokes ML-knn on the data after dimension reduction, while MSE is directly applicable to the original high dimensional data.

\begin{multicols}{2}

\begin{table}[H]
\caption{Prediction performances (\%) and CPU seconds of BR \cite{MLreview2}, ML-KNN \cite{MLKNN}, MDDM \cite{MDDM} and MSE on Yahoo. Prec-precision, Rec-recall, F1-F1 score, Acc-accuracy}
\begin{center}
\begin{tabular}{|c|l|*{5}{c}|}
\hline
& Methods & Prec & Rec & F1 & Acc & CPU sec. \\
\hline
\multirow{4}{*}{\begin{sideways}Arts\end{sideways}}&BR 	&$76$	&$25$	&$26$	&$24$	&$46.8$ \\
&ML-knn 	&$62$	&$7$	&$25$	&$6$	&$77.6$ \\
&MDDM 	&$68$	&$6$	&$21$	&$5$	&$37.4$ \\
&MSE 	&$35$	&$40$	&$31$	&$28$	&$11.7$ \\
\hline
\multirow{4}{*}{\begin{sideways}Education\end{sideways}}&BR 	&$69$	&$27$	&$28$	&$26$	&$50.1$ \\
&ML-knn 	&$58$	&$6$	&$31$	&$5$	&$99.8$ \\
&MDDM 	&$59$	&$5$	&$26$	&$5$	&$45.2$ \\
&MSE 	&$41$	&$35$	&$32$	&$29$	&$12.6$ \\
\hline
\multirow{4}{*}{\begin{sideways}Recreation\end{sideways}}&BR 	&$84$	&$23$	&$23$	&$22$	&$53.2$ \\
&ML-knn 	&$70$	&$9$	&$23$	&$8$	&$112$ \\
&MDDM 	&$66$	&$7$	&$18$	&$6$	&$41.9$ \\
&MSE 	&$41$	&$49$	&$36$	&$30$	&$19.1$ \\
\hline
\multirow{4}{*}{\begin{sideways}Science\end{sideways}}&BR	&$79$	&$19$	&$19$	&$19$	&$84.9$ \\
&ML-knn 	&$59$	&$4$	&$20$	&$4$	&$139$ \\
&MDDM 	&$66$	&$4$	&$19$	&$4$	&$53.0$ \\
&MSE 	&$31$	&$39$	&$29$	&$26$	&$20.1$ \\
\hline
\multirow{4}{*}{\begin{sideways}Business\end{sideways}}&BR	&$87$	&$74$	&$76$	&$71$	&$28.9$ \\
&ML-knn 	&$68$	&$9$	&$70$	&$8$	&$93.2$ \\
&MDDM 	&$66$	&$7$	&$69$	&$7$	&$42.7$ \\
&MSE 	&$84$	&$82$	&$78$	&$78$	&$13.5$ \\
\hline
\end{tabular}
\end{center}
\label{Table:yahoo}
\end{table}

\begin{table}[H]
\caption{Prediction performances (\%) and CPU seconds of BR \cite{MLreview2}, ML-KNN \cite{MLKNN}, MDDM \cite{MDDM} and MSE on 8 datasets. Prec-precision, Rec-recall, F1-F1 score, Acc-accuracy}
\begin{center}
\begin{tabular}{|c|l|*{5}{c}|}
\hline
& Methods & Prec & Rec & F1 & Acc & CPU sec. \\
\hline
\multirow{4}{*}{\begin{sideways}Mediamill\end{sideways}} &BR 	&$69$	&$35$	&$43$	&$33$	&$120141$ \\
&ML-knn &$41$	&$6$	&$54$	&$5$	&$5713$ \\
&MDDM &$36$	&$5$	&$53$	&$4$	&$48237$ \\
&MSE &$58$	&$78$	&$53$	&$37$	&$1155$ \\
\hline
\multirow{4}{*}{\begin{sideways}Enron\end{sideways}}&BR 	&$51$	&$28$	&$35$	&$24$	&$77.1$ \\
&ML-knn &$51$	&$7$	&$46$	&$5$	&$527$ \\
&MDDM &$50$	&$8$	&$49$	&$7$	&$29$ \\
&MSE &$44$	&$50$	&$40$	&$28$	&$271$ \\
\hline
\multirow{4}{*}{\begin{sideways}Medical\end{sideways}}&BR 	&$2$	&$26$	&$5$	&$2$	&$4.88$ \\
&ML-knn &$75$	&$7$	&$48$	&$6$	&$22.8$ \\
&MDDM &$74$	&$3$	&$30$	&$2$	&$32.3$ \\
&MSE &$36$	&$90$	&$45$	&$26$	&$7.5$ \\
\hline
\multirow{4}{*}{\begin{sideways}Slashdot\end{sideways}}&BR 	&$11$	&$22$	&$14$	&$10$	&$140$ \\
&ML-knn &$71$	&$10$	&$31$	&$8$	&$708$ \\
&MDDM &$39$	&$1$	&$4$	&$1$	&$114$ \\
&MSE &$38$	&$61$	&$37$	&$27$	&$175$ \\
\hline
\multirow{4}{*}{\begin{sideways}Scene\end{sideways}}&BR	&$55$	&$67$	&$66$	&$63$	&$4.19$ \\
&ML-knn &$78$	&$62$	&$69$	&$54$	&$14.3$ \\
&MDDM &$75$	&$64$	&$69$	&$53$	&$7.59$ \\
&MSE &$61$	&$85$	&$70$	&$68$	&$3.62$ \\
\hline
\multirow{4}{*}{\begin{sideways}Emotions\end{sideways}}&BR 	&$55$	&$53$	&$51$	&$42$	&$0.68$ \\
&ML-knn &$68$	&$28$	&$41$	&$22$	&$0.66$ \\
&MDDM &$54$	&$28$	&$41$	&$22$	&$0.66$ \\
&MSE &$40$	&$100$	&$52$	&$37$	&$0.01$ \\
\hline
\multirow{4}{*}{\begin{sideways}Genbase\end{sideways}}&BR	&$5$	&$39$	&$9$	&$5$	&$1.99$ \\
&ML-knn &$100$	&$50$	&$92$	&$50$	&$9.38$ \\
&MDDM &$98$	&$51$	&$92$	&$51$	&$6.09$ \\
&MSE &$83$	&$96$	&$86$	&$70$	&$8.62$ \\
\hline
\multirow{4}{*}{\begin{sideways}Corel5k\end{sideways}} &BR  & $2$ & $20$ & $4$ & $2$ & $2240$ \\
&ML-knn 	&$62$	&$1$	&$3$	&$0.9$	&$2106$ \\
&MDDM 	&$62$	&$1$	&$7$	&$1$	&$458$ \\
&MSE 	&$9$	&$11$	&$8$	&$5$	&$1054$ \\
\hline
\end{tabular}
\end{center}
\label{Table:exp}
\end{table}

\end{multicols}

In the comparison of performance via the four metrics, the F1 score and accuracy of MSE outperform those of other methods on most datasets. Moreover, MSE has smaller gaps between precision and recall on different tasks than other methods, and this implies it is robust to the imbalance between positive and negative samples. Note in multi-label prediction, only large values of all four metrics are sufficient to indicate the success of the prediction, while the combination of some large valued metrics and some small valued ones are always caused by the imbalance of the samples. Therefore, MSE provides better prediction performance than other methods on most datasets.

\begin{figure*}[h!]\label{fig:Phase750}
\begin{center}\vspace{-3mm}
\subfigure{
 \includegraphics[width=0.45\linewidth]{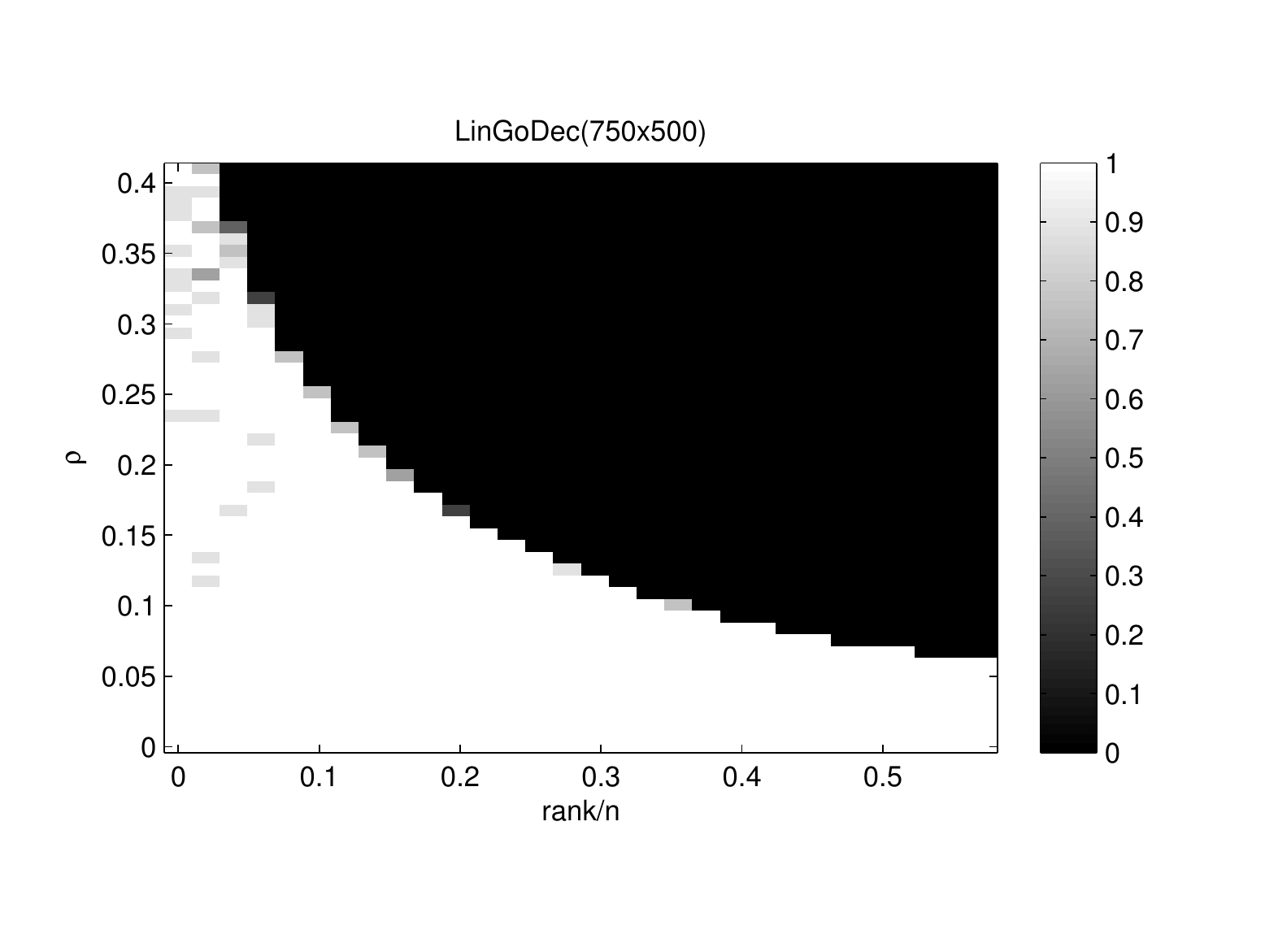}
}
\subfigure{
 \includegraphics[width=0.45\linewidth]{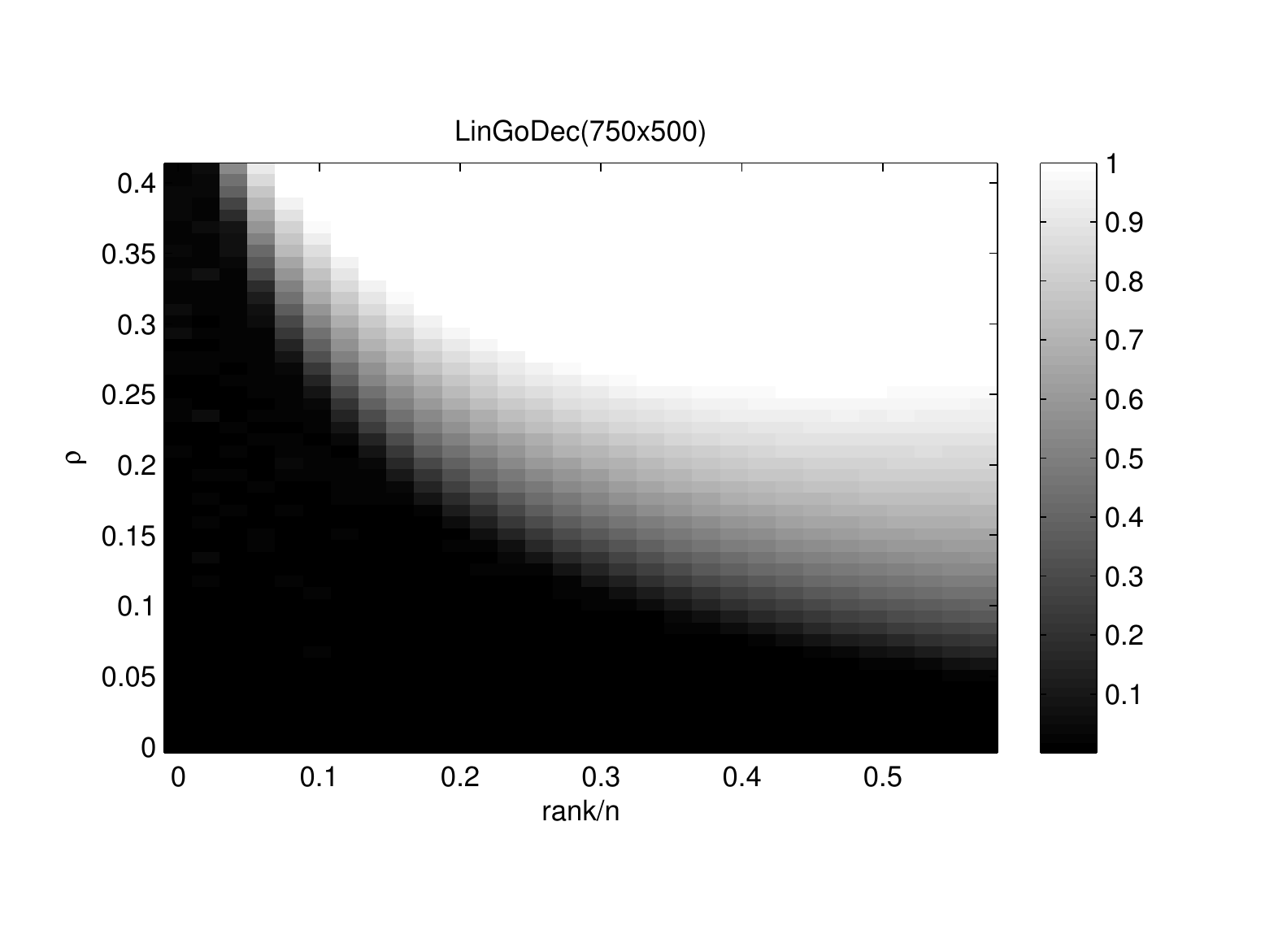}
}
\end{center}\vspace{-5mm}
\caption{Phase diagram (left) and corresponding CPU seconds (right) for LinGoDec on $750\times 750$ matrices. Low-rank weight matrix $W$ is of size $750\times 500$, and is generated by $W=UV$, where entries of $U$ and $V$ are sampled from $\mathcal N(0,1/750)$ and $\mathcal N(0,1/750)$, respectively. Features of items in $Z$ is sampled from $\mathcal N(0,1/750)$. Entries of sparse anomaly $S$ are sampled as $1$ or $-1$ with probability $\rho/2$ and $0$ with probability $1-\rho$. Noise $G$ has entries sampled from $\mathcal N(0,10^{-3})$. On the $50\times 30$ grid of sparsity-rank/n plane, $10$ trials are performed for each $(\rho,r)$ pair. $W$ is said to be successfully recovered if its rel. err.$\leq 10^{-2}$. The phase diagram shows the successful recovery rate for each $(\rho,r)$ pair.}
\end{figure*}

\subsection{LinGoDec on Synthetic Data}

Since most public available dataset for recommendation system rarely fulfill our demands for the training data in LinGoDec, we justify LinGoDec on synthetic data. Specifically, the rating matrix $X$ is generated by $WZ^T+S+G$. The weight matrix of linear functions $W$ is generated as the product of two Gaussian matrices. Entries in both the item feature matrix $Z$ and noise matrix $G$ are generated by i.i.d. Gaussian distribution. The sparse part has a Bernoulli model generated support set on which $\pm 1$ values are randomly assigned. 

We show the phase diagram and the corresponding time cost in Figure \ref{fig:Phase750}. It could be seem that LinGoDec has a slightly larger region (the white region) for successful recovery than both GreBsmo and robust PCA \cite{RobustPCA}. This is because side-information, i.e., the features of items, is utilized in LinGoDec. Moreover, the time cost of LinGoDec is still small due to the closed-form update of $W$ and BRP based acceleration. 

Therefore, LinGoDec is capable to achieve the scoring functions of users, which cannot be learned by previous matrix completion based methods, and is effective to rule out the advertising effects in user ratings. Its fast speed makes it very efficient when applied to practical systems.


\small{
\bibliographystyle{IEEEtran}
\bibliography{GoDecArxiv}
}

\newpage

\section*{Appendix I: Analysis of GoDec}

We theoretically analyze the convergence of GoDec. The objective value (decomposition error) $\|X-L-S\|_F^2$ monotonically decreases and converges to a local minimum. Since the updating of $L$ and $S$ in GoDec is equivalent to alternatively projecting $L$ or $S$ onto two smooth manifolds, we use the framework proposed in \cite{AlternatingP} to prove the asymptotical property and linear convergence of $L$ and $S$. The asymptotic and convergence speeds are mainly determined by the angle between the two manifolds. We discuss how $L$, $S$ and $G$ influence the speeds via influencing the cosine of the angle. The analyses show the convergence of GoDec is robust to the noise $G$.

In particular, we first prove that the objective value $\|X-L-S\|_F^2$ (decomposition error) converges to a local minimum. Then we demonstrate the asymptotic properties of GoDec and prove that the solutions $L$ and $S$ respectively converge to local optimums with linear rate less than $1$. The influence of $L$, $S$ and $G$ to the asymptotic/convergence speeds is analyzed. The speeds are slowed down by augmenting the magnitude of noise part $\|G\|_F^2$. However, the convergence still holds unless $\|G\|_F^2\gg\|L\|_F^2$ or $\|G\|_F^2\gg\|S\|_F^2$.

We have the following theorem about the convergence of the objective value $\|X-L-S\|_F^2$ in (\ref{E:ls_app}).
\begin{theorem}\label{T:ls_convergence}
{\rm (\textbf{Convergence of objective value})}. The alternative optimization (\ref{E:raw_ls}) produces a sequence of $\|X-L-S\|_F^2$ that converges to a local minimum.
\end{theorem}
\begin{proof}
Let the objective value $\|X-L-S\|_F^2$ after solving the two subproblems in (\ref{E:raw_ls}) be $E_t^1$ and $E_t^2$, respectively, in the $t^{th}$ iteration. On the one hand, we have
\begin{equation}
E_t^1=\|X-L_t-S_{t-1}\|_F^2, E_t^2=\|X-L_t-S_t\|_F^2.
\end{equation}
The global optimality of $S_t$ yields $E_t^1\geq E_t^2$. On the other hand,
\begin{equation}
E_t^2=\|X-L_t-S_t\|_F^2, E_{t+1}^1=\|X-L_{t+1}-S_t\|_F^2.
\end{equation}
The global optimality of $L_{t+1}$ yields $E_t^2\geq E_{t+1}^1$. Therefore, the objective values (decomposition errors) $\|X-L-S\|_F^2$ keep decreasing throughout GoDec (\ref{E:raw_ls}):
\begin{equation}\label{E:converge}
E_1^1\geq E_1^2\geq E_2^1\geq\cdots\geq E_t^1\geq E_t^2\geq E_{t+1}^1\geq\cdots
\end{equation}
Since the objective of (\ref{E:ls_app}) is monotonically decreasing and the constraints are satisfied all the time, (\ref{E:raw_ls}) produces a sequence of objective values that converge to a local minimum. This completes the proof.
\end{proof}

The asymptotic property and the linear convergence of $L$ and $S$ in GoDec are demonstrated based on the framework proposed in \cite{AlternatingP}. We firstly consider $L$. From a different prospective, GoDec algorithm shown in (\ref{E:raw_lss}) is equivalent to iteratively projecting $L$ onto one manifold $\mathcal M$ and then onto another manifold $\mathcal N$. This kind of optimization method is the so called ``alternating projections on manifolds''. To see this, in (\ref{E:raw_lss}), by substituting $S_{t}$ into the next updating of $L_{t+1}$, we have:
\begin{equation}
L_{t+1}=\mathcal P_{\mathcal M}\left(X-\mathcal P_\Omega\left(X-L_t\right)\right)=\mathcal P_{\mathcal M}\mathcal P_{\mathcal N}\left(L_t\right),
\end{equation}
Both $\mathcal M$ and $\mathcal N$ are two $C^k$-manifolds around a point $\overline L\in\mathcal M\cap\mathcal N$:
\begin{equation}\label{E:manifoldMN}
\left\{
  \begin{array}{ll}
    \mathcal M=\left\{H\in\mathbb R^{m\times n}:{\rm rank}\left(H\right)=r\right\},\\
    \mathcal N=\left\{X-\mathcal P_\Omega\left(X-H\right):H\in\mathbb R^{m\times n}\right\}.
  \end{array}
\right.
\end{equation}
According to the above definitions, any point $L\in\mathcal M\cap\mathcal N$ satisfies:
\begin{align}
&L=\mathcal P_{\mathcal M\cap\mathcal N}\left(L\right)\Rightarrow \\
&L=X-\mathcal P_\Omega\left(X-L\right),{\rm rank}\left(L\right)=r.
\end{align}
Thus any point $L\in\mathcal M\cap\mathcal N$ is a local solution of $L$ in (\ref{E:ls_app}).

We define the angle between two manifolds $\mathcal M$ and $\mathcal N$ at point $L$ as the angle between the corresponding tangent spaces $T_{\mathcal M}(L)$ and $T_{\mathcal N}(L)$. The angle is between $0$ and $\pi/2$ with cosine:
\begin{equation}\label{E:tangentangle}
c\left(\mathcal M,\mathcal N,L\right)=c\left(T_{\mathcal M}(L),T_{\mathcal N}(L)\right).
\end{equation}
In addition, if $\mathbb S$ is the unit sphere in $\mathbb R^{m\times n}$, the angle between two subspaces $M$ and $N$ in $\mathbb R^{m\times n}$ is defined as the angle between $0$ and $\pi/2$ with cosine:
\begin{equation}
\begin{array}{ll}
\notag c\left(M,N\right)=\max\left\{\langle x,y \rangle:x\in\mathbb S\cap M\cap\left(M\cap N\right)^\bot,\right.\\
\left.~~~~~~~~~~~~~~~~~~~~~~~~~~~~~~~~~~~~~~~~~~~y\in\mathbb S\cap N\cap\left(M\cap N\right)^\bot\right\}.
\end{array}
\end{equation}

We give the following proposition about the angle between two subspaces $M$ and $N$:
\begin{proposition}\label{P:SetCal}
Following the above definition of the angle between two subspaces $M$ and $N$, we have
\begin{equation}
\begin{array}{ll}
\notag c\left(M,N\right)=\max\left\{\langle x,y \rangle:x\in\mathbb S\cap M\cap N^\bot,\right.\\
\left.~~~~~~~~~~~~~~~~~~~~~~~~~~~~~~~~~~~~~~~~~~~y\in\mathbb S\cap N\cap M^\bot\right\}.
\end{array}
\end{equation}
\end{proposition}

The angle between $\mathcal M$ and $\mathcal N$ is used in the asymptotical property and the linear convergence rate of ``alternating projections on manifolds'' algorithms.

\begin{theorem}\label{T:AsymptPro}
{\rm (\textbf{Asymptotic property} \cite{AlternatingP})}. Let $\mathcal M$ and $\mathcal N$ be two transverse $C^2$-manifolds around a point $\overline L\in\mathcal M\cap\mathcal N$. Then
\begin{equation}
\notag\limsup_{L\rightarrow\overline L,L\notin\mathcal M\cap\mathcal N}\frac{\left\|\mathcal P_{\mathcal M}\mathcal P_{\mathcal N}\left(L\right)-\mathcal P_{\mathcal M\cap\mathcal N}\left(L\right)\right\|}{\left\|L-\mathcal P_{\mathcal M\cap\mathcal N}\left(L\right)\right\|}\leq c\left(\mathcal M,\mathcal N,\overline L\right).
\end{equation}
A refinement of the above argument is
\begin{equation}
\notag\limsup_{L\rightarrow\overline L,L\notin\mathcal M\cap\mathcal N}\frac{\left\|\left(\mathcal P_{\mathcal M}\mathcal P_{\mathcal N}\right)^n\left(L\right)-\mathcal P_{\mathcal M\cap\mathcal N}\left(L\right)\right\|}{\left\|L-\mathcal P_{\mathcal M\cap\mathcal N}\left(L\right)\right\|}\leq c^{2n-1}
\end{equation}
for $n=1,2,...$ and $c=c\left(\mathcal M,\mathcal N,\overline L\right)$.
\end{theorem}

\begin{theorem}\label{T:LinearCov}
{\rm (\textbf{Linear convergence of variables} \cite{AlternatingP})}. In $\mathbb R^{m\times n}$, let $\mathcal M$ and $\mathcal N$ be two transverse manifolds around a point $\overline L\in\mathcal M\cap\mathcal N$. If the initial point $L_0\in\mathbb R^{m\times n}$ is close to $\overline L$, then the method of alternating projections
\begin{equation}
\notag L_{t+1}=\mathcal P_{\mathcal M}\mathcal P_{\mathcal N}\left(L_t\right),(t=0,1,2,...)
\end{equation}
is well-defined, and the distance $d_{\mathcal M\cap\mathcal N}(L_t)$ from the iterate $L_t$ to the intersection $\mathcal M\cap\mathcal N$ decreases Q-linearly to zero. More precisely, given any constant $c$ strictly larger than the cosine of the angle of the intersection between the manifolds, $\c(\mathcal M,\mathcal N,\overline L)$, if $L_0$ is close to $\overline L$, then the iterates satisfy
\begin{equation}
\notag d_{\mathcal M\cap\mathcal N}(L_{t+1})\leq c\cdot d_{\mathcal M\cap\mathcal N}(L_t),(t=0,1,2,...)
\end{equation}
Furthermore, $L_t$ converges linearly to some point $L^*\in\mathcal M\cap\mathcal N$, i.e., for some constant $\alpha>0$,
\begin{equation}
\notag \left\|L_t-L^*\right\|\leq\alpha c^t,(t=0,1,2,...).
\end{equation}
\end{theorem}

Since GoDec algorithm can be written as the form of alternating projections on two manifolds $\mathcal M$ and $\mathcal N$ given in (\ref{E:manifoldMN}) and they satisfy the assumptions of Theorem \ref{T:AsymptPro} and Theorem \ref{T:LinearCov}, $L$ in GoDec converges to a local optimum with linear rate. Similarly, we can prove the linear convergence of $S$.

Since cosine $c(\mathcal M,\mathcal N,\overline L)$ in Theorem \ref{T:AsymptPro} and Theorem \ref{T:LinearCov} determines the asymptotic and convergence speeds of the algorithm. We discuss how $L$, $S$ and $G$ influence the asymptotic and convergence speeds via analyzing the relationship between $L$, $S$, $G$ and $c(\mathcal M,\mathcal N,\overline L)$.

\begin{theorem}\label{T:ACspeed}
{\rm (\textbf{Asymptotic and convergence speed})}. In GoDec, the asymptotical improvement and the linear convergence of $L$ and $S$ stated in Theorem \ref{T:AsymptPro} and Theorem \ref{T:LinearCov} will be slowed by augmenting
\begin{align}
\notag&{\rm For~}L:\frac{\left\|\Delta_L\right\|_F}{\left\|L+\Delta_L\right\|_F},\Delta_L=\left(S+G\right)-\mathcal P_{\Omega}\left(S+G\right),\\
\notag&{\rm For~}S:\frac{\left\|\Delta_S\right\|_F}{\left\|S+\Delta_S\right\|_F},\Delta_S=\left(L+G\right)-\mathcal P_{\mathcal M}\left(L+G\right).
\end{align}
However, the asymptotical improvement and the linear convergence will not be harmed and is robust to the noise $G$ unless when $\|G\|_F\gg\|S\|_F$ and $\|G\|_F\gg\|L\|_F$, which lead the two terms increasing to $1$.
\end{theorem}
\begin{proof}
GoDec approximately decomposes a matrix $X=L+S+G$ into the low-rank part $L$ and the sparse part $S$. According to the above analysis, GoDec is equivalent to alternating projections of $L$ on $\mathcal M$ and $\mathcal N$, which are given in (\ref{E:manifoldMN}). According to Theorem \ref{T:AsymptPro} and Theorem \ref{T:LinearCov}, smaller $c(\mathcal M,\mathcal N,\overline L)$ produces faster asymptotic and convergence speeds, while $c(\mathcal M,\mathcal N,\overline L)=1$ is possible to make $L$ and $S$ stopping converging. Below we discuss how $L$, $S$ and $G$ influence $c(\mathcal M,\mathcal N,\overline L)$ and further influence the asymptotic and convergence speeds of GeDec.

According to (\ref{E:tangentangle}), we have
\begin{equation}
c\left(\mathcal M,\mathcal N,\overline L\right)=c\left(T_{\mathcal M}(\overline L),T_{\mathcal N}(\overline L)\right).
\end{equation}
Substituting the equation given in Proposition \ref{P:SetCal} into the right-hand side of the above equation yields
\begin{equation}\label{E:cosMN}
\begin{array}{ll}
c\left(\mathcal M,\mathcal N,\overline L\right)=\max\left\{\langle x,y \rangle:x\in\mathbb S\cap T_{\mathcal M}(\overline L)\cap N_{\mathcal N}(\overline L),\right.\\
\left.~~~~~~~~~~~~~~~~~~~~~~~~~y\in\mathbb S\cap T_{\mathcal N}(\overline L)\cap N_{\mathcal M}(\overline L)\right\}.\\
\end{array}
\end{equation}
The normal spaces of manifolds $\mathcal M$ and $\mathcal N$ on point $\overline L$ is respectively given by
\begin{equation}\label{E:normalMN}
\begin{array}{ll}
&N_{\mathcal M}(\overline L)=\left\{y\in\mathbb R^{m\times n}:u_i^Tyv_j=0,\overline L=UDV^T\right\},\\
&N_{\mathcal N}(\overline L)=\left\{X-\mathcal P_\Omega\left(X-\overline L\right)\right\},
\end{array}
\end{equation}
where $\overline L=UDV^T$ represents the eigenvalue decomposition of $\overline L$, $U=[u_1,...,u_r]$ and $V=[v_1,...,v_r]$. Assume $X=\overline L+\overline S+\overline G$, wherein $\overline G$ is the noise corresponding to $\overline L$, we have
\begin{align}
\notag&\overline L=X-\left(\overline S+\overline G\right),\\
\notag&\hat L=X-\mathcal P_\Omega\left(\overline S+\overline G\right),\Rightarrow\\
&\hat L=\overline L+\left[\left(\overline S+\overline G\right)-\mathcal P_\Omega\left(\overline S+\overline G\right)\right]=\overline L+\Delta.
\end{align}
Thus the normal space of manifold $\mathcal N$ is
\begin{equation}\label{E:normalN}
N_{\mathcal N}(\overline L)=\left\{\overline L+\Delta\right\}.
\end{equation}
Since the tangent space is the complement space of the normal space, by using the normal space of $\mathcal M$ in (\ref{E:normalMN}) and the normal space of $\mathcal N$ given in (\ref{E:normalN}), we can verify
\begin{equation}\label{E:subsetMN}
N_{\mathcal N}(\overline L)\subseteq T_{\mathcal M}(\overline L), N_{\mathcal M}(\overline L)\subseteq T_{\mathcal N}(\overline L).
\end{equation}
By substituting the above results into (\ref{E:cosMN}), we obtain
\begin{equation}
\begin{array}{ll}
c\left(\mathcal M,\mathcal N,\overline L\right)=\max\left\{\langle x,y \rangle:x\in\mathbb S\cap N_{\mathcal N}(\overline L),\right.\\
\left.~~~~~~~~~~~~~~~~~~~~~~~~~~~~~~~~~~~~~~~~~~~~~~~y\in\mathbb S\cap N_{\mathcal M}(\overline L)\right\}.\\
\end{array}
\end{equation}
Hence we have
\begin{align}
\notag\langle x,y \rangle&={\rm tr}\left(VDU^Ty+\Delta^Ty\right)\\
&={\rm tr}\left(DU^TyV\right)+{\rm tr}\left(\Delta^Ty\right)={\rm tr}\left(\Delta^Ty\right).
\end{align}
The last equivalence is due to $u_i^Tyv_j=0$ in (\ref{E:normalMN}). Thus
\begin{equation}
c\left(\mathcal M,\mathcal N,\overline L\right)=\max\left\{\langle x,y \rangle\right\}\leq\max\left\{\langle D_\Delta,D_y \rangle\right\},
\end{equation}
where the diagonal entries of $D_\Delta$ and $D_y$ are composed by eigenvalues of $\Delta$ and $y$, respectively. The last inequality is obtained by considering the case when $x$ and $y$ have identical left and right singular vectors. Because $\overline L+\Delta,y\in\mathbb S$ infers $\|\overline L+\Delta\|_F^2=\|y\|_F^2=1$, we have
\begin{align}
\notag c\left(\mathcal M,\mathcal N,\overline L\right)&\leq\max\left\{\langle D_\Delta,D_y \rangle\right\}\\
&\leq\left\|D_\Delta\right\|_F\left\|D_y\right\|_F\leq\left\|D_\Delta\right\|_F.
\end{align}
Since $c$ in Theorem \ref{T:LinearCov} can be selected as any constant that is strictly larger than $c\left(\mathcal M,\mathcal N,\overline L\right)\leq\left\|D_\Delta\right\|_F$, we can choose $c=c\left(\mathcal M,\mathcal N,\overline L\right)+\Delta_c\leq\left\|D_\Delta\right\|_F$. In Theorem \ref{T:AsymptPro}, the cosine $c\left(\mathcal M,\mathcal N,\overline L\right)$ is directly used.

Therefore, the asymptotic and convergence speeds of $L$ will be slowed by augmenting $\|\Delta\|_F$, and vice versa. However, the asymptotical improvement and the linear convergence will not be jeopardized unless $\|\Delta\|_F=1$. For general $L+\Delta$ that is not normalized onto the sphere $\mathbb S$, $\|\Delta\|_F$ should be replaced by $\|\Delta\|_F/\|L+\Delta\|_F$.

For the variable $S$, we can obtain an analogous result via an analysis in a similar style as above. For general $L+\Delta$ without normalization, the asymptotic/convergence speed of $S$ will be slowed by augmenting $\|\Delta\|_F/\|S+\Delta\|_F$, and vice versa, wherein
\begin{equation}
\Delta=\left(L+G\right)-\mathcal P_{\mathcal M}\left(L+G\right).
\end{equation}
The asymptotical improvement and the linear convergence will not be jeopardized unless $\|\Delta\|_F/\|S+\Delta\|_F=1$.

This completes the proof.
\end{proof}

Theorem \ref{T:ACspeed} reveals the influence of the low-rank part $L$, the sparse part $S$ and the noise part $G$ to the asymptotic/convergence speeds of $L$ and $S$ in GoDec. Both $\Delta_L$ and $\Delta_S$ are the element-wise hard thresholding error of $S+G$ and the singular value hard thresholding error of $L+G$, respectively. Large errors will slow the asymptotic and convergence speeds of GoDec. Since $S-\mathcal P_\Omega(S)=0$ and $L-\mathcal P_{\mathcal M}(L)=0$, the noise part $G$ in $\Delta_L$ and $\Delta_S$ can be interpreted as the perturbations to $S$ and $L$ and deviates the two errors from $0$. Thus noise $G$ with large magnitude will decelerate the asymptotical improvement and the linear convergence, but it will not ruin the convergence unless $\|G\|_F\gg\|S\|_F$ or $\|G\|_F\gg\|L\|_F$. Therefore, GoDec is robust to the additive noise in $X$ and is able to find the approximated $L+S$ decomposition when noise $G$ is not overwhelming.

\newpage

\section*{Appendix II: Approximation Error Bound of BRP}

\section{Approximation error bounds}

We analyze the error bounds of the BRP based low-rank approximation (\ref{E:lr_app}) and its power scheme modification (\ref{E:mlr_app}).

The SVD of an $m\times n$ (w.l.o.g, $m\geq n$) matrix $X$ takes the form:
\begin{equation}
X=U\Lambda V^T=U_1\Lambda_1V_1^T+U_2\Lambda_2V_2^T,
\end{equation}
where $\Lambda_1$ is an $r\times r$ diagonal matrix which diagonal elements are the first largest $r$ singular values, $U_1$ and $V_1$ are the corresponding singular vectors, $\Lambda_2$, $U_2$ and $V_2$ forms the rest part of SVD. Assume that $r$ is the target rank, $A_1$ and $A_2$ have $r+p$ columns for oversampling. We consider the spectral norm of the approximation error $E$ for (\ref{E:lr_app}):
\begin{align}
\notag\|X-L\|&=\left\|X-Y_1\left(A_2^TY_1\right)^{-1}Y_2^T\right\|\\
&=\left\|\left[I-XA_1\left(A_2^TXA_1\right)^{-1}A_2^T\right]X\right\|.
\end{align}
The unitary invariance of the spectral norm leads to
\begin{align}\label{E:unblock}
\notag \|X-L\|=\left\|U^T\left[I-XA_1\left(A_2^TXA_1\right)^{-1}A_2^T\right]X\right\|\\
~~~~~~~~~~=\left\|\Lambda\left[I-V^TA_1\left(A_2^TXA_1\right)^{-1}A_2^TU\Lambda\right]\right\|.
\end{align}

In low-rank approximation, the left random projection matrix $A_2$ is built from the left random projection $Y_1=XA_1$, and then the right random projection matrix $A_1$ is built from the left random projection $Y_2=X^TA_2$. Thus $A_2=Y_1=XA_1=U\Lambda V^TA_1$ and $A_1=Y_2=X^TA_2=X^TXA_1=V\Lambda^2V^TA_1$. Hence the approximation error given in (\ref{E:unblock}) has the following form:
\begin{equation}\label{E:newunblock}
\left\|\Lambda\left[I-\Lambda^2V^TA_1\left(A_1^TV\Lambda^4V^TA_1\right)^{-1}A_1^TV\Lambda^2\right]\right\|.
\end{equation}

The following Theorem \ref{T:deterministicBound} gives the bound for the spectral norm of the deterministic error $\|X-L\|$.

\begin{theorem}\label{T:deterministicBound}
{\rm \textbf{(Deterministic error bound)}} Given an $m\times n\left(m\geq n\right)$ real matrix $X$ with singular value decomposition $X=U\Lambda V^T=U_1\Lambda_1 V_1^T+U_2\Lambda_2 V_2^T$, and chosen a target rank $r\leq n-1$ and an $n\times(r+p)$ ($p\geq2$) standard Gaussian matrix $A_1$, the BRP based low-rank approximation (\ref{E:lr_app}) approximates $X$ with the error upper bounded by
\begin{equation}
\notag\|X-L\|^2\leq\left\|\Lambda_2^2\left(V_2^TA_1\right)(V_1^TA_1)^\dagger\Lambda_1^{-1}\right\|^2+\left\|\Lambda_2\right\|^2.
\end{equation}
\end{theorem}

See Section \ref{S:proof} for the proof of Theorem \ref{T:deterministicBound}.

If the singular values of $X$ decay fast, the first term in the deterministic error bound will be very small. The last term is the rank-$r$ SVD approximation error. Therefore, the BRP based low-rank approximation (\ref{E:lr_app}) is nearly optimal.

\begin{theorem}\label{T:deterministicBoundPower}
{\rm \textbf{(Deterministic error bound, power scheme)}} Frame the hypotheses of Theorem \ref{T:deterministicBound}, the power scheme modification (\ref{E:mlr_app}) approximates $X$ with the error upper bounded by
\begin{align}
\notag\|X-L\|^2\leq&\left(\left\|\Lambda_2^{2(2q+1)}\left(V_2^TA_1\right)\left(V_1^TA_1\right)^\dagger\Lambda_1^{-(2q+1)}\right\|^2\right.\\
\notag &\left.+\left\|\Lambda_2^{2q+1}\right\|^2\right)^{1/(2q+1)}.
\end{align}
\end{theorem}

See Section \ref{S:proof} for the proof of Theorem \ref{T:deterministicBoundPower}.

If the singular values of $X$ decay slowly, the error produced by the power scheme modification (\ref{E:mlr_app}) is less than the BRP based low-rank approximation (\ref{E:lr_app}) and decreasing with the increasing of $q$.

The average error bound of BRP based low-rank approximation is obtained by analyzing the statistical properties of the random matrices that appear in the deterministic error bound in Theorem \ref{T:deterministicBound}.

\begin{theorem}\label{T:AverageErrorBound}
{\rm \textbf{(Average error bound)}} Frame the hypotheses of Theorem \ref{T:deterministicBound},
\begin{align}
\notag\mathbb E\|X-L\|\leq&\left(\sqrt{\frac{1}{p-1}\sum\limits_{i=1}^r\frac{\lambda_{r+1}^2}{\lambda_i^2}}+1\right)|\lambda_{r+1}|\\
\notag &+\frac{{\rm e}\sqrt{r+p}}{p}\sqrt{\sum\limits_{i=r+1}^n\frac{\lambda_i^2}{\lambda_r^2}}.
\end{align}
\end{theorem}

See Section \ref{S:proof} for the proof of Theorem \ref{T:AverageErrorBound}.

The average error bound will approach to the SVD approximation error $|\lambda_{r+1}|$ if $|\lambda_{r+1}|\ll|\lambda_{i:i=1,\cdots,r}|$ and $|\lambda_{r}|\gg|\lambda_{i:i=r+1,\cdots,n}|$.

The average error bound for the power scheme modification is then obtained from the result of Theorem \ref{T:AverageErrorBound}.

\begin{theorem}\label{T:AverageErrorBoundPower}
{\rm (\textbf{Average error bound, power scheme})} Frame the hypotheses of Theorem \ref{T:deterministicBound}, the power scheme modification (\ref{E:mlr_app}) approximates $X$ with the expected error upper bounded by
\begin{align}
\notag E\|X-L\|\leq&\left[\left(\sqrt{\frac{1}{p-1}\sum\limits_{i=1}^r\frac{\lambda_{r+1}^{2(2q+1)}}{\lambda_i^{2(2q+1)}}}+1\right)|\lambda_{r+1}^{2q+1}|\right.\\
\notag &\left.+\frac{{\rm e}\sqrt{r+p}}{p}\sqrt{\sum\limits_{i=r+1}^n\frac{\lambda_i^{2(2q+1)}}{\lambda_r^{2(2q+1)}}}\right]^{1/(2q+1)}.
\end{align}
\end{theorem}

See Section \ref{S:proof} for the proof of Theorem \ref{T:AverageErrorBoundPower}.

Compared the average error bounds of the BRP based low-rank approximation with its power scheme modification, the latter produces less error than the former, and the error can be further decreased by increasing $q$.

The deviation bound for the spectral norm of the approximation error can be obtained by analyzing the deviation bound of $\left\|\Lambda_2^2\left(V_2^TA_1\right)(V_1^TA_1)^\dagger\Lambda_1^{-1}\right\|$ in the deterministic error bound and by applying the concentration inequality for Lipschitz functions of a Gaussian matrix.

\begin{theorem}\label{T:DeviationErrorBound}
{\rm (\textbf{Deviation bound})} Frame the hypotheses of Theorem \ref{T:deterministicBound}. Assume that $p\geq4$. For all $u,t\geq1$, it holds that
\begin{align}
\notag \left\|X-L\right\|\leq&\left(1+t\sqrt{\frac{12r}{p}}\left(\sum\limits_{i=1}^r\lambda_i^{-1}\right)^{\frac{1}{2}}+\frac{{\rm e}\sqrt{r+p}}{p+1}\cdot\right.\\
\notag &\left.tu\lambda_r^{-1}\right)\lambda_{r+1}^2+\frac{{\rm e}\sqrt{r+p}}{p+1}\cdot t\lambda_r^{-1}\left(\sum\limits_{i=r+1}^n\lambda_i^2\right)^{\frac{1}{2}}.
\end{align}
except with probability ${\rm e}^{-u^2/2}+4t^{-p}+t^{-(p+1)}$.
\end{theorem}

See Section \ref{S:proof} for the proof of Theorem \ref{T:DeviationErrorBound}.

\subsection{Proofs of error bounds}
\label{S:proof}

\subsubsection{Proof of Theorem \ref{T:deterministicBound}}

The following lemma and propositions from \cite{RandomSVD} will be used in the proof.

\begin{lemma}\label{L:conjugate}
Suppose that $M\succeq0$. For every $A$, the matrix $A^TMA\succeq0$. In particular,
\begin{equation}
M\preceq N~~\Rightarrow~~A^TMA\preceq A^TNA.
\end{equation}
\end{lemma}

\begin{proposition}\label{P:Range}
Suppose ${\rm range}(N)\subset{\rm range}(M)$. Then, for each matrix $A$, it holds that $\|\mathcal P_NA\|\leq\|\mathcal P_MA\|$ and that $\|(I-\mathcal P_M)A\|\leq\|(I-\mathcal P_N)A\|$.
\end{proposition}

\begin{proposition}\label{P:inversePurtubation}
Suppose that $M\succeq0$. Then
\begin{equation}
I-\left(I+M\right)^{-1}\preceq M.
\end{equation}
\end{proposition}

\begin{proposition}\label{P:blockNorm}
We have $\|M\|\leq\|A\|+\|C\|$ for each partitioned positive semidefinite matrix
\begin{equation}
M=\left[
    \begin{array}{cc}
      A & B \\
      B^T & C \\
    \end{array}
  \right].
\end{equation}
\end{proposition}

The proof of Theorem \ref{T:deterministicBound} is given below.

\begin{proof}
Since an orthogonal projector projects a given matrix to the range (column space) of a matrix $M$ is defined as $\mathcal P_M=M(M^TM)^{-1}M^T$, the deterministic error (\ref{E:newunblock}) can be written as
\begin{equation}\label{E:errorproj}
\|E\|=\left\|\Lambda\left(I-\mathcal P_M\right)\right\|,~M=\Lambda^2V^TA_1.
\end{equation}

By applying Proposition \ref{P:Range} to the error (\ref{E:errorproj}), because ${\rm range}(M(V_1^TA_1)^\dagger\Lambda_1^{-2})\subset{\rm range}(M)$, we have
\begin{equation}\label{E:projNM}
\|E\|=\left\|\Lambda\left(I-\mathcal P_M\right)\right\|\leq\left\|\Lambda\left(I-\mathcal P_N\right)\right\|,
\end{equation}
where
\begin{equation}\label{E:IsubPN}
N=\left[
    \begin{array}{c}
      \Lambda_1^2V_1^TA_1 \\
      \Lambda_2^2V_2^TA_1 \\
    \end{array}
  \right]
(V_1^TA_1)^\dagger\Lambda_1^{-2}=
\left[
  \begin{array}{c}
    I \\
    H \\
  \end{array}
\right]
.
\end{equation}
Thus $\left(I-\mathcal P_N\right)$ can be written as
\begin{equation}\label{E:IsubPNBlock}
\notag I-\mathcal P_N=
\left[
  \begin{array}{cc}
    I-\left(I+H^TH\right)^{-1} & -\left(I+H^TH\right)^{-1}H^T \\
    -H\left(I+H^TH\right)^{-1} & I-H\left(I+H^TH\right)^{-1}H^T \\
  \end{array}
\right]
\end{equation}

For the top-left block in (\ref{E:IsubPNBlock}), Proposition \ref{P:inversePurtubation} leads to $I-\left(I+H^TH\right)^{-1}\preceq H^TH$. For the bottom-right block in (\ref{E:IsubPNBlock}), Lemma \ref{L:conjugate} leads to $I-H\left(I+H^TH\right)^{-1}H^T\preceq I$. Therefore,
\begin{equation}
\notag I-\mathcal P_N\preceq\left[
                       \begin{array}{cc}
                         H^TH & -\left(I+H^TH\right)^{-1}H^T \\
                         -H\left(I+H^TH\right)^{-1} & I \\
                       \end{array}
                     \right]
\end{equation}

By applying Lemma \ref{L:conjugate}, we have
\begin{align}
\notag&\Lambda\left(I-\mathcal P_N\right)\Lambda\preceq\\
\notag&\left[
                                                  \begin{array}{cc}
                                                    \Lambda_1^TH^TH\Lambda_1 & -\Lambda_1^T\left(I+H^TH\right)^{-1}H^T\Lambda_2 \\
                                                    -\Lambda_2^TH\left(I+H^TH\right)^{-1}\Lambda_1 & \Lambda_2^T\Lambda_2 \\
                                                  \end{array}
                                                \right]
\end{align}

According to Proposition \ref{P:blockNorm}, the spectral norm of $\Lambda(I-\mathcal P_N)$ is bounded by
\begin{align}\label{E:rawBound}
\notag&\left\|\Lambda\left(I-\mathcal P_N\right)\right\|^2=\left\|\Lambda\left(I-\mathcal P_N\right)\Lambda\right\|\\
&\leq\left\|\Lambda_2^2\left(V_2^TA_1\right)(V_1^TA_1)^\dagger\Lambda_1^{-1}\right\|^2+\left\|\Lambda_2\right\|^2.
\end{align}

By substituting (\ref{E:rawBound}) into (\ref{E:projNM}), we obtain the deterministic error bound. This completes the proof.
\end{proof}

\subsubsection{Proof of Theorem \ref{T:deterministicBoundPower}}

The following proposition from \cite{RandomSVD} will be used in the proof.

\begin{proposition}\label{P:PowerNorm}
Let $\mathcal P$ be an orthogonal projector, and let $A$ be a matrix. For each nonnegative $q$,
\begin{equation}
\|\mathcal PA\|\leq\left\|\mathcal P\left(AA^T\right)^qA\right\|^{1/\left(2q+1\right)}.
\end{equation}
\end{proposition}

The proof of Theorem \ref{T:deterministicBoundPower} is given below.

\begin{proof}
The power scheme modification (\ref{E:mlr_app}) applies the BRP based low-rank approximation (\ref{E:lr_app}) to $\tilde X=(XX^T)^qX=U\Lambda^{2q+1}V^T$ rather than $X$. In this case, the approximation error is
\begin{equation}
\|\tilde X-\tilde L\|=\left\|\Lambda^{2q+1}\left(I-\mathcal P_M\right)\right\|,~M=\Lambda^{2(2q+1)}V^TA_1.
\end{equation}
According to Theorem \ref{T:deterministicBound}, the error is upper bounded by
\begin{align}\label{E:unblockPower}
\notag&\left\|\tilde X-\tilde L\right\|^2\leq\\
&\left\|\Lambda_2^{2(2q+1)}\left(V_2^TA_1\right)(V_1^TA_1)^\dagger\Lambda_1^{-(2q+1)}\right\|^2+\left\|\Lambda_2^{2q+1}\right\|^2.
\end{align}
The deterministic error bound for the power scheme modification is obtained by applying Proposition \ref{P:PowerNorm} to (\ref{E:unblockPower}). This completes the proof.
\end{proof}

\subsubsection{Proof of Theorem \ref{T:AverageErrorBound}}

The following propositions from \cite{RandomSVD} will be used in the proof.

\begin{proposition}\label{P:SGT}
Fix matrices $S$, $T$, and draw a standard Gaussian matrix $G$. Then it holds that
\begin{equation}
\mathbb E\left\|SGT^T\right\|\leq\|S\|\|T\|_F+\|S\|_F\|T\|.
\end{equation}
\end{proposition}

\begin{proposition}\label{P:pesudoinvGaussian}
Draw an $r\times(r+p)$ standard Gaussian matrix $G$ with $p\geq 2$. Then it holds that
\begin{align}
\mathbb E\|G^\dagger\|_F^2=\frac{r}{p-1},
\mathbb E\|G^\dagger\|\leq\frac{{\rm e}\sqrt{r+p}}{p}.
\end{align}
\end{proposition}

The proof of Theorem \ref{T:AverageErrorBound} is given below.

\begin{proof}
The distribution of a standard Gaussian matrix is rotational invariant. Since 1) $A_1$ is a standard Gaussian matrix and 2) $V$ is an orthogonal matrix, $V^TA_1$ is a standard Gaussian matrix, and its disjoint submatrices $V_1^TA_1$ and $V_2^TA_1$ are standard Gaussian matrices as well.

Theorem \ref{T:deterministicBound} and the H\"{o}lder's inequality imply that
\begin{align}\label{E:EXsubL}
\notag \mathbb E\|X-L\|&\leq\mathbb E\left(\left\|\Lambda_2^2\left(V_2^TA_1\right)(V_1^TA_1)^\dagger\Lambda_1^{-1}\right\|^2+\|\Lambda_2\|^2\right)^{1/2}\\
&\leq\mathbb E\left\|\Lambda_2^2\left(V_2^TA_1\right)(V_1^TA_1)^\dagger\Lambda_1^{-1}\right\|+\|\Lambda_2\|.
\end{align}
We condition on $V_1^TA_1$ and apply Proposition \ref{P:SGT} to bound the expectation w.r.t. $V_2^TA_1$, i.e.,
\begin{align}\label{E:ExAA}
\notag &E\left\|\Lambda_2^2\left(V_2^TA_1\right)(V_1^TA_1)^\dagger\Lambda_1^{-1}\right\|\\
\notag &\leq\mathbb E\left(\left\|\Lambda_2^2\right\|\left\|(V_1^TA_1)^\dagger\Lambda_1^{-1}\right\|_F+\left\|\Lambda_2^2\right\|_F\left\|(V_1^TA_1)^\dagger\Lambda_1^{-1}\right\|\right)\\
\notag&\leq\left\|\Lambda_2^2\right\|\left(\mathbb E\left\|(V_1^TA_1)^\dagger\Lambda_1^{-1}\right\|_F^2\right)^{1/2}+\\
&\left\|\Lambda_2^2\right\|_F\cdot\mathbb E\left\|(V_1^TA_1)^\dagger\right\|\cdot\left\|\Lambda_1^{-1}\right\|.
\end{align}
The Frobenius norm of $(V_1^TA_1)^\dagger\Lambda_1^{-1}$ can be calculated as
\begin{align}
\notag \left\|(V_1^TA_1)^\dagger\Lambda_1^{-1}\right\|_F^2&={\rm trace}\left[\Lambda_1^{-1}\left((V_1^TA_1)^\dagger\right)^T(V_1^TA_1)^\dagger\Lambda_1^{-1}\right]\\
\notag&={\rm trace}\left[\left(\left(\Lambda_1V_1^TA_1\right)\left(\Lambda_1V_1^TA_1\right)^T\right)^{-1}\right].
\end{align}
Since 1) $V_1^TA_1$ is a standard Gaussian matrix and 2) $\Lambda_1$ is a diagonal matrix, each column of $\Lambda_1V_1^TA_1$ follows $r$-variate Gaussian distribution $\mathcal N_r(\textbf{0},\Lambda_1^2)$. Thus the random matrix $\left(\left(\Lambda_1V_1^TA_1\right)\left(\Lambda_1V_1^TA_1\right)^T\right)^{-1}$ follows the inverted Wishart distribution $\mathcal W^{-1}_r(\Lambda_1^{-2},r+p)$. According to the expectation of inverted Wishart distribution \cite{AspectsStats}, we have
\begin{align}
\notag &\mathbb E\left\|(V_1^TA_1)^\dagger\Lambda_1^{-1}\right\|_F^2\\
\notag &=\mathbb E~{\rm trace}\left[\left(\left(\Lambda_1V_1^TA_1\right)\left(\Lambda_1V_1^TA_1\right)^T\right)^{-1}\right]\\
\notag &={\rm trace}~\mathbb E\left[\left(\left(\Lambda_1V_1^TA_1\right)\left(\Lambda_1V_1^TA_1\right)^T\right)^{-1}\right]\\
&=\frac{1}{p-1}\sum\limits_{i=1}^r\lambda_i^{-2}.
\end{align}
We apply Proposition \ref{P:pesudoinvGaussian} to the standard Gaussian matrix $V_1^TA_1$ and obtain
\begin{equation}
\mathbb E\left\|(V_1^TA_1)^\dagger\right\|\leq\frac{{\rm e}\sqrt{r+p}}{p}.
\end{equation}
Therefore, (\ref{E:ExAA}) can be further derived as
\begin{align}\label{E:ExAAfinal}
\notag E&\left\|\Lambda_2^2\left(V_2^TA_1\right)(V_1^TA_1)^\dagger\Lambda_1^{-1}\right\|\\
\notag&\leq\lambda_{r+1}^2\cdot\sqrt{\frac{1}{p-1}\sum\limits_{i=1}^r\lambda_i^{-2}}+\sqrt{\sum\limits_{i=r+1}^n\lambda_i^2}\cdot\frac{{\rm e}\sqrt{r+p}}{p}\cdot|\lambda_r^{-1}|\\
&=|\lambda_{r+1}|\sqrt{\frac{1}{p-1}\sum\limits_{i=1}^r\frac{\lambda_{r+1}^2}{\lambda_i^2}}+\frac{{\rm e}\sqrt{r+p}}{p}\sqrt{\sum\limits_{i=r+1}^n\frac{\lambda_i^2}{\lambda_r^2}}.
\end{align}
By substituting (\ref{E:ExAAfinal}) into (\ref{E:EXsubL}), we obtain the average error bound
\begin{align}
\notag\mathbb E\|X-L\|\leq&\left(\sqrt{\frac{1}{p-1}\sum\limits_{i=1}^r\frac{\lambda_{r+1}^2}{\lambda_i^2}}+1\right)|\lambda_{r+1}|+\\
&\frac{{\rm e}\sqrt{r+p}}{p}\sqrt{\sum\limits_{i=r+1}^n\frac{\lambda_i^2}{\lambda_r^2}}.
\end{align}
This completes the proof.
\end{proof}

\subsubsection{Proof of Theorem \ref{T:AverageErrorBoundPower}}

The proof of Theorem \ref{T:AverageErrorBoundPower} is given below.

\begin{proof}
By using H\"{o}lder's inequality and Theorem \ref{T:deterministicBoundPower}, we have
\begin{align}\label{E:powerXsubL}
\notag\mathbb E\left\|X-L\right\|&\leq\left(\mathbb E\left\|X-L\right\|^{2q+1}\right)^{1/(2q+1)}\\
&\leq\left(\mathbb E\left\|\tilde X-\tilde L\right\|\right)^{1/(2q+1)}.
\end{align}
We apply Theorem \ref{T:AverageErrorBound} to $\tilde X$ and $\tilde L$ and obtain the bound of $\mathbb E\|\tilde X-\tilde L\|$, noting that $\lambda_i(\tilde X)=\lambda_i(X)^{2q+1}$.
\begin{align}\label{E:tildeXsubL}
\notag \mathbb E\left\|\tilde X-\tilde L\right\|=&\left(\sqrt{\frac{1}{p-1}\sum\limits_{i=1}^r\frac{\lambda_{r+1}^{2(2q+1)}}{\lambda_i^{2(2q+1)}}}+1\right)|\lambda_{r+1}^{2q+1}|+\\
&\frac{{\rm e}\sqrt{r+p}}{p}\sqrt{\sum\limits_{i=r+1}^n\frac{\lambda_i^{2(2q+1)}}{\lambda_r^{2(2q+1)}}}.
\end{align}
By substituting (\ref{E:tildeXsubL}) into (\ref{E:powerXsubL}), we obtain the average error bound of the power scheme modification shown in Theorem \ref{T:AverageErrorBoundPower}. This completes the proof.
\end{proof}

\subsubsection{Proof of Theorem \ref{T:DeviationErrorBound}}

The following propositions from \cite{RandomSVD} will be used in the proof.

\begin{proposition}\label{P:LipschitzConcentration}
Suppose that $h$ is a Lipschitz function on matrices:
\begin{equation}
\left|h(X)-h(Y)\right|\leq L\|X-F\|_F~~for~all~X,Y.
\end{equation}
Draw a standard Gaussian matrix $G$. Then
\begin{equation}
\Pr\left\{h(G)\geq\mathbb Eh(G)+Lt\right\}\leq{\rm e}^{-t^2/2}.
\end{equation}
\end{proposition}

\begin{proposition}\label{P:GaussianNormDeviation}
Let $G$ be a $r\times(r+p)$ standard Gaussian matrix where $p\geq4$. For all $t\geq1$,
\begin{align}
\notag&\Pr\left\{\left\|G^\dagger\right\|_F\geq\sqrt{\frac{12r}{p}}\cdot t\right\}\leq4t^{-p}~~{\rm and}\\
&\Pr\left\{\left\|G^\dagger\right\|\geq\frac{{\rm e}\sqrt{r+p}}{p+1}\cdot t\right\}\leq t^{-(p+1)}.
\end{align}
\end{proposition}

The proof of Theorem \ref{T:DeviationErrorBound} is given below.

\begin{proof}
According to the deterministic error bound in Theorem \ref{T:deterministicBound}, we study the deviation of $\left\|\Lambda_2^2\left(V_2^TA_1\right)\left(V_1^TA_1\right)^\dagger\Lambda_1^{-1}\right\|$. Consider the Lipschitz function $h(X)=\left\|\Lambda_2^2X\left(V_1^TA_1\right)^\dagger\Lambda_1^{-1}\right\|$, its Lipschitz constant $L$ can be estimated by using the triangle inequality:
\begin{align}
\notag &\left|h(X)-h(Y)\right| \leq\left\|\Lambda_2^2\left(X-Y\right)\left(V_1^TA_1\right)^\dagger\Lambda_1^{-1}\right\|\\
\notag &\leq\left\|\Lambda_2^2\right\|\left\|X-Y\right\|\left\|\left(V_1^TA_1\right)^\dagger\right\|\left\|\Lambda_1^{-1}\right\|\\
&\leq\left\|\Lambda_2^2\right\|\left\|\left(V_1^TA_1\right)^\dagger\right\|\left\|\Lambda_1^{-1}\right\|\left\|X-Y\right\|_F.
\end{align}
Hence the Lipschitz constant satisfies $L\leq\left\|\Lambda_2^2\right\|\left\|\left(V_1^TA_1\right)^\dagger\right\|\left\|\Lambda_1^{-1}\right\|$. We condition on $V_1^TA_1$ and then Proposition \ref{P:SGT} implies that
\begin{align}
\notag\mathbb E\left[h\left(V_2^TA_1\right)\left|\right.V_1^TA_1\right]\leq&\left\|\Lambda_2^2\right\|\left\|\left(V_1^TA_1\right)^\dagger\right\|_F\left\|\Lambda_1^{-1}\right\|_F+\\
\notag&\left\|\Lambda_2^2\right\|_F\left\|\left(V_1^TA_1\right)^\dagger\right\|\left\|\Lambda_1^{-1}\right\|.
\end{align}
We define an event $T$ as
\begin{align}
\notag&T=\left\{\left\|\left(V_1^TA_1\right)^\dagger\right\|_F\leq\sqrt{\frac{12r}{p}}\cdot t~~{\rm and}~~\right.\\
&\left.~~~~~~~~~\left\|\left(V_1^TA_1\right)^\dagger\right\|\leq\frac{{\rm e}\sqrt{r+p}}{p+1}\cdot t\right\}.
\end{align}
According to Proposition \ref{P:GaussianNormDeviation}, the event $T$ happens except with probability
\begin{equation}
\Pr\left\{\overline{T}\right\}\leq4t^{-p}+t^{-(p+1)}.
\end{equation}
Applying Proposition \ref{P:LipschitzConcentration} to the function $h\left(V_2^TA_1\right)$, given the event $T$, we have
\begin{align}
\notag\Pr&\left\{\left\|\Lambda_2^2\left(V_2^TA_1\right)\left(V_1^TA_1\right)^\dagger\Lambda_1^{-1}\right\|>\right.\\
\notag&\left\|\Lambda_2^2\right\|\left\|\left(V_1^TA_1\right)^\dagger\right\|_F\left\|\Lambda_1^{-1}\right\|_F+\\
\notag&{\left\|\Lambda_2^2\right\|_F\left\|\left(V_1^TA_1\right)^\dagger\right\|\left\|\Lambda_1^{-1}\right\|+}\\
&\left.{\left\|\Lambda_2^2\right\|\left\|\left(V_1^TA_1\right)^\dagger\right\|\left\|\Lambda_1^{-1}\right\|\cdot u}\mid T\right\}\leq{\rm e}^{-u^2/2}.
\end{align}
According to the definition of the event $T$ and the probability of $\overline{T}$, we obtain
\begin{align}
\notag\Pr&\left\{\left\|\Lambda_2^2\left(V_2^TA_1\right)\left(V_1^TA_1\right)^\dagger\Lambda_1^{-1}\right\|>\right.\\
\notag&\left\|\Lambda_2^2\right\|\left\|\Lambda_1^{-1}\right\|_F\sqrt{\frac{12r}{p}}\cdot t+\left\|\Lambda_2^2\right\|_F\left\|\Lambda_1^{-1}\right\|\frac{{\rm e}\sqrt{r+p}}{p+1}\cdot t\\
\notag&\left.+\left\|\Lambda_2^2\right\|\left\|\Lambda_1^{-1}\right\|\frac{{\rm e}\sqrt{r+p}}{p+1}\cdot tu\right\}\leq\\
\notag&{\rm e}^{-u^2/2}+4t^{-p}+t^{-(p+1)}.
\end{align}
Therefore,
\small
\begin{align}
\notag\Pr&\left\{\left\|\Lambda_2^2\left(V_2^TA_1\right)\left(V_1^TA_1\right)^\dagger\Lambda_1^{-1}\right\|+\left\|\Lambda_2\right\|>\right.\\
\notag&\left.\left(1+t\sqrt{\frac{12r}{p}}\left(\sum\limits_{i=1}^r\lambda_i^{-1}\right)^{1/2}+\frac{{\rm e}\sqrt{r+p}}{p+1}\cdot tu\lambda_r^{-1}\right)\lambda_{r+1}^2+\right.\\
\notag&\left.\frac{{\rm e}\sqrt{r+p}}{p+1}\cdot t\lambda_r^{-1}\left(\sum\limits_{i=r+1}^n\lambda_i^2\right)^{1/2}\right\}\leq\\
&{\rm e}^{-u^2/2}+4t^{-p}+t^{-(p+1)}.
\end{align}
\normalsize
Since Theorem \ref{T:deterministicBound} implies $\left\|X-L\right\|\leq\left\|\Lambda_2^2\left(V_2^TA_1\right)\left(V_1^TA_1\right)^\dagger\Lambda_1^{-1}\right\|+\left\|\Lambda_2\right\|$,
we obtain the deviation bound in Theorem \ref{T:DeviationErrorBound}. This completes the proof.
\end{proof}

\newpage

\section*{Appendix III: Analysis of GreB}

It is not direct to analyze the theoretical guarantee of GreB due to its combination of alternating minimization and greedy forward selection. Hence, we consider analyzing its convergence behavior by leveraging the results from GECO \cite{GECO} analysis. This is reasonable because they share the same objective function yet different optimization variables. In particular, the risk function in GECO is $R(A)=R(A(\lambda))=f(\lambda)$, where $A=\sum_{i}\lambda_iU_iV_i$. It can be seen that the variable $A$ in GECO is able to be written as $A=UV$ without any loss of generality. Therefore, for the same selection of $R(A)$, we can compare the objective value of GECO and GreB at arbitrary step of their algorithm. This results in the following theorem.
\begin{theorem}
Assume $R(A)$ is a $\beta$-smooth function according to GECO \cite{GECO} and $\epsilon>0$, and $F(U,V)=R(UV)$ is the objective function of GreB. Given a rank constraint $r$ to $A$ and a tolerance parameter $\tau\in\left[\right.0,1\left.\right)$. Let $A^*=U^*V^*$ is the solution of GreB. Then for all matrices $A=UV$ with
\begin{equation}
\|UV\|_{tr}^2\leq\frac{\epsilon(r+1)(1-\tau)^2}{2\beta}
\end{equation}
we have $F(U^*,V^*)\leq F(U,V)+\epsilon$.
\end{theorem}
\begin{proof}
According to Lemma 3 in GECO \cite{GECO}, let $\epsilon_i=f(\lambda^{(i)})-f(\bar\lambda)$, where $\lambda^{(i)}$ is the value of $\lambda$ at the beginning of iteration $i$ and $\bar\lambda$ fulfills $f(\lambda)>f(\bar\lambda)$, we have
\begin{equation}\label{equ:GECOiequ}
f(\lambda^{(i)})-\min\limits_\eta f(\lambda^{(i)}+\eta e^{u,v})\geq \frac{\epsilon_i^2(1-\tau)^2}{2\beta\|A\|_{tr}^2}.
\end{equation}
At the end of iteration $i$, the objective value of GreB equals $R(UV)$, while GECO optimizes $\lambda$ over the support of ${\rm span}(U)\times{\rm span}(V)$ (i.e., optimizes $S$ when fixing $U$ and $V$). We use the same notation $\cdot^{(i)}$ to denote the variable in iteration $i$. This yields
\begin{equation}\label{equ:ith}
\begin{array}{ll}
F(U^{(i)},V^{(i)})=&R(U^{(i)}V^{(i)})\geq \\
&\min\limits_S R(U^{(i)}SV^{(i)})=f(\lambda^{(i)}).
\end{array}
\end{equation}
At the beginning of iteration $i+1$, both GECO and GreB computes the direction $(u,v)$ along which the object declines fastest. However, GECO adds both $u$ and $v$ to the ranges of $U$ and $V$, while GreB only adds $v$ to $V$ and then optimizes $U$ when fixing $V$. Because the range of $U$ in GreB is optimized rather than previously fixed, we have
\begin{equation}\label{equ:i1th}
\begin{array}{ll}
F(U^{(i+1)},V^{(i+1)})=&\min\limits_U F(U,[V^{(i+1)};v])\leq\\
&\min\limits_\eta f(\lambda^{(i)}+\eta e^{u,v}).
\end{array}
\end{equation}
Plug (\ref{equ:ith}) and (\ref{equ:i1th}) into (\ref{equ:GECOiequ}), we gain a similar result:
\begin{equation}
F(U,V)-\min\limits_U F(U,[V;v])\geq \frac{\epsilon_i^2(1-\tau)^2}{2\beta\|A\|_{tr}^2}.
\end{equation}
Following the analysis after Lemma 3 in GECO \cite{GECO}, we can immediately obtain the results of the theorem.
\end{proof}
The theorem states that GreB solution is at least close to optimum as GECO. Note when sparse $S$ is alternatively optimized with $UV$ in GreB scheme, such as GreBcom, the theorem can still holds. This is because after optimizing $S$ in each iteration of GreBcom, we have $\mathcal P_{\Omega^C}(S+UV)=0$, which enforces the objective function $\|M-UV-S\|_F^2$ degenerates to that of GECO, which is $\|P_{\Omega}(M-UV)\|_F^2$.

\end{document}